\documentclass[11pt]{article}
\usepackage{xcolor}
\usepackage{multirow}
\usepackage{booktabs}
\usepackage[utf8]{inputenc}  
\usepackage{supertabular}
\usepackage{threeparttable}
\usepackage{amsmath,amsfonts}
\usepackage{algorithmic}
\usepackage{algorithm}
\usepackage{array}
\usepackage[caption=false,font=normalsize,labelfont=sf,textfont=sf]{subfig}
\usepackage{textcomp}
\usepackage{stfloats}
\usepackage{url}
\usepackage{pdflscape}
\usepackage{verbatim}
\usepackage{graphicx}
\usepackage{cite}
\usepackage[hidelinks]{hyperref}
\begin{document}

\title{PMMA: The Polytechnique Montréal Mobility Aids Dataset}

\author{Qingwu Liu, 
Nicolas Saunier, 
and Guillaume-Alexandre Bilodeau
\thanks{Qingwu Liu and Nicolas Saunier are from Civil, Geological and Mining Engineering Department, Guillaume-Alexandre Bilodeau is from Computer Engineering and Software Engineering Department, Polytechnique Montréal, 2500 Chem. de Polytechnique, Montréal, Canada, H3T 1J4. Contact: qingwu.liu@polymtl.ca, nicolas.saunier@polymtl.ca and guillaume-alexandre.bilodeau@polymtl.ca}
}


\markboth{Journal of IEEE Transactions on Intelligent Transportation Systems}%
{Shell \MakeLowercase{\textit{et al}}: A Sample Article Using IEEEtran.cls for IEEE Journals}


\maketitle

\begin{abstract}
This study introduces a new object detection dataset of pedestrians using mobility aids, named PMMA. The dataset was collected in an outdoor environment, where volunteers used wheelchairs, canes, and walkers, resulting in nine categories of pedestrians: pedestrians, cane users, two types of walker users, whether walking or resting, five types of wheelchair users, including wheelchair users, people pushing empty wheelchairs, and three types of users pushing occupied wheelchairs, including the entire pushing group, the pusher and the person seated on the wheelchair. 
To establish a benchmark, seven object detection models (Faster R-CNN, CenterNet, YOLOX, DETR, Deformable DETR, DINO, and RT-DETR) and three tracking algorithms (ByteTrack, BOT-SORT, and OC-SORT) were implemented under the MMDetection framework. Experimental results show that YOLOX, Deformable DETR, and Faster R-CNN achieve the best detection performance, while the differences among the three trackers are relatively small. The PMMA dataset is publicly available at: \url{https://doi.org/10.5683/SP3/XJPQUG} and the video processing and model training code is available at: \url{https://github.com/DatasetPMMA/PMMA}.
\end{abstract}


\section{Introduction}

Cameras have become ubiquitous in traffic environments, serving as essential tools for capturing and analyzing road users to better understand their behaviors and anticipate their movements. In both computer vision and intelligent transportation systems, object detection and tracking are fundamental tasks. 
Existing publicly available object detection and tracking datasets classify road users into broad categories such as pedestrians, cyclists, cars and trucks. However, few datasets provide finer-grained distinctions among pedestrians, particularly pedestrians who use mobility aids such as wheelchairs, walkers, or canes. Although individuals using mobility aids represent a small fraction of road users, they are more vulnerable than other groups. This underscores the need for dedicated datasets that specifically identify mobility aid users and enable a thorough evaluation of state-of-the-art (SOTA) detection and tracking algorithms.
Such efforts are crucial for advancing inclusive and safety-focused intelligent transportation systems. 

Image-based object detection and tracking methods have seen significant progress, driven not only by the advancement of computational power, the increase of GPU resources, but also by the evolution of deep learning models, such as convolutional neural networks (CNN)~\cite{krizhevsky2012imagenet} and Transformers~\cite{vaswani2017attention}, the availability of large-scale annotated datasets, and improved training techniques. These advancements have enabled SOTA models to achieve strong performance on public datasets such as COCO~\cite{lin2014microsoft}, KITTI~\cite{Geiger2012CVPR}, and Cityscapes~\cite{cordts2016cityscapes}. However, due to the limited category granularity in existing datasets, how these methods perform when faced with multiple visually similar categories remains an open question.

\begin{figure}[!t]
    \centering
    \includegraphics[width=\linewidth]{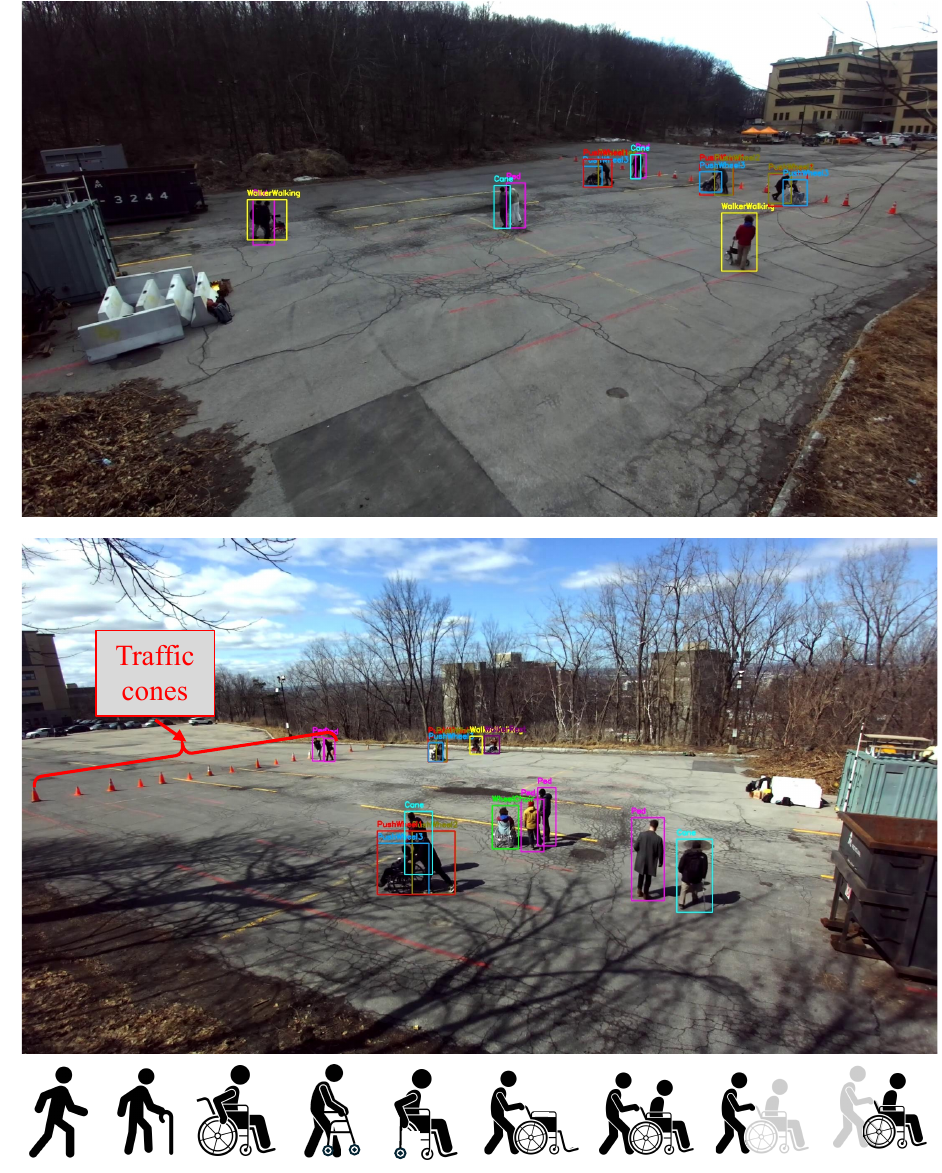}
    \caption{Example frames of our Mobility Aids Dataset. There are nine categories, with detailed descriptions in Table~\ref{tab:pedestrian_categories} and icons adapted from~\cite{icons_1, icons_2}}
    \label{fig:example_of_our_dataset}
\end{figure}

Therefore, we introduce in this paper our mobility aids dataset, which contains over 28,000 annotated images with a resolution of 2208 $\times$ 1242 pixels and a maximum frame rate of 15 frames per second (fps). We collected the dataset using a ZED 2 stereo camera in an outdoor parking lot of Polytechnique Montréal, located in Montréal, Québec, Canada. Three videos were annotated from two points of view 
with nine categories of pedestrians using computer vision annotation tools (CVAT)~\cite{CVAT_ai_Corporation_Computer_Vision_Annotation_2023}, as visualized in Figure~\ref{fig:example_of_our_dataset}.  We compare our dataset with existing object detection datasets in Table~\ref{tab:dataset_comparison}. 

To evaluate the performance of detectors and trackers in our dataset, we benchmark seven detectors and three trackers on our dataset. We further discuss their performance and the shortcomings. Finally, we summarize the most challenging categories and the limitations of those methods on our dataset.

The main contributions of this study are summarized as follows: 
\begin{enumerate}
    \item We constructed a dataset containing nine categories of pedestrians with mobility aids, with annotations provided for all images and occlusion information recorded for each annotated instance;
    \item We applied seven object detection methods and three tracking methods on the dataset; and
    \item We analyzed the experimental results and highlighted the main challenges associated with the detection and tracking of pedestrians using mobility aids.
\end{enumerate}

The remainder of this paper is organized as follows. Section~\ref{sec:RelatedWork} presents the related work. Section~\ref{sec:PMMA_Dataset} introduces the details of our dataset.  Section~\ref{sec:Experiments} presents the experimental protocol and evaluation metrics. Section~\ref{sec:ResultsAndDiscussions} presents the results and the discussion, and section~\ref{sec:Conclusion} concludes the paper.

\section{Related Work}
\label{sec:RelatedWork}
In this section, prior datasets and data collection paradigms relevant to our work are reviewed. We begin by discussing autonomous-driving datasets collected from vehicle-mounted sensors, which provide rich annotations but remain limited by car-centric visibility and short sequence durations. We then review fixed-camera and multi-camera surveillance datasets that offer high-density pedestrian tracking from elevated viewpoints. Finally, we highlight datasets that, similar to ours, focus on fine-grained pedestrian sub-categories, particularly those involving mobility aids.

Several widely used datasets in the autonomous driving domain have been collected using vehicle-mounted sensors. Notable examples include KITTI~\cite{Geiger2012CVPR}, Waymo Open Dataset~\cite{waymo}, SemanticKITTI~\cite{sematicKITTI}, JAAD~\cite{rasouli2017jaad}, nuScenes~\cite{nuscence}, and Cityscapes~\cite{cordts2016cityscapes}. 
These datasets provide high-quality sensor data, such as LiDAR and RGB images, collected as vehicles drive through urban environments. 
The low camera position and occlusions by other vehicles limit visibility, and most of the data is captured in multiple short-duration segments, typically lasting only a few seconds. Even with longer sequences, detailed traffic analysis remains challenging due to the inherent limitations of the vehicle perspective. In addition, although these datasets are useful for tasks, such as object detection and semantic segmentation in road scenes, they mainly focus on general object categories, such as pedestrians, cars, and cyclists, without distinguishing between pedestrians with or without mobility aids.


In crowded public spaces, several datasets focus on fixed-camera surveillance or multi-camera scenarios. MOT20~\cite{mot20} offers high-density pedestrian tracking sequences  from viewpoints at lamppost or building height. The A9-Dataset~\cite{cress2022a9} provides high-resolution multi-modal data collected from roadside sensor infrastructure along a three kilometers stretch of the A9 autobahn near Munich, Germany. It includes precision-timestamped camera and LiDAR frames captured from overhead gantry bridges, with objects annotated using 3D bounding boxes. WILDTRACK~\cite{WILDTRACK} and CityFlow~\cite{CityFlow} are multi-view datasets that allow for 3D localization and multi-camera tracking. LUMPI~\cite{LUMPI} is another multi-perspective dataset designed for analyzing pedestrian behavior across viewpoints. While these datasets provide multi-angle coverage and dense pedestrian annotations, they do not include detailed pedestrian categories either.

Similar to our dataset, a few datasets categorize pedestrians into multiple subclasses to enable fine-grained analysis.
SynPoses~\cite{nie2022synposes} proposes a framework for generating high-quality synthetic human data with diverse and complex poses, addressing the limited pose variability in existing real-world datasets.
RAP~\cite{li2018richly} is a large-scale pedestrian dataset designed for person retrieval in real surveillance environments, featuring rich attribute annotations and identity labels that enable both attribute-based and image-based person re-identification. The dataset mainly focuses on appearance-related attributes, such as clothing styles and hair characteristics, while mobility-aid users are not explicitly considered.
A few datasets explicitly consider categories related to mobility aids. For instance, the dataset proposed by
Sonia Dávila-Soberón et al.~\cite{davila2025novel} specifically focuses on wheelchair users, cane users, and the mobility aids themselves. However, this dataset is not publicly available, which limits its accessibility for broader research purposes. Another dataset~\cite{kollmitz2019deep} targets mobility aid users in an indoor hospital environment, providing valuable data for healthcare-related applications but lacking outdoor scene variability and typical viewpoints. In addition, Zhang et al.~\cite{zhang2021x} introduced a synthetic dataset in a simulation environment called X-World, which includes objects such as wheelchairs and canes. While useful for controlled experiments, simulation-based data often lacks the complexity and diversity of real-world environments.

To the best of our knowledge, there is currently no publicly available dataset that captures in real-world outdoor conditions people using diverse mobility aids, such as wheelchairs, walkers and canes, while also supporting tasks like object detection and multi-object tracking (MOT). This highlights a gap in existing datasets for developing and evaluating vision-based systems that aim to understand and assist mobility-impaired individuals in dynamic, open outdoor environments.

\begin{landscape}
\begin{table*}[]
\centering
\caption{Comparison of the PMMA dataset statistics with existing benchmarks}
\label{tab:dataset_comparison}
\begin{tabular}{lcccccc}
\toprule
\textbf{Dataset} & \textbf{View} & \textbf{Images} & \textbf{Categories with persons} & \textbf{Outdoor} & \textbf{Real world} & \textbf{Mobility Aids} \\
\midrule
\textbf{Vehicle-perspective} & & & & & & \\
KITTI~\cite{Geiger2012CVPR}        & ego   & 15k   & 3 & \checkmark & \checkmark & $\times$ \\
Waymo~\cite{waymo}        & ego   & 1M    & 2 & \checkmark & \checkmark & $\times$ \\
Cityscapes~\cite{cordts2016cityscapes}   & ego   & 25k   & 2 & \checkmark & \checkmark & $\times$ \\
JAAD~\cite{rasouli2017jaad}       & ego   & 346k  & 1 & \checkmark & \checkmark & $\times$ \\
nuScenes~\cite{nuscence} & ego & 1.4M & 2 & \checkmark & \checkmark & $\times$ \\
\midrule
\textbf{Multi-perspective} & & & & & & \\
WILDTRACK~\cite{WILDTRACK}    & surv+ego  & 215k  & 1 & \checkmark & \checkmark & $\times$ \\
CityFlow~\cite{CityFlow}     & surv+surv & 118k  & 2 & \checkmark & \checkmark & $\times$ \\
A9-Dataset~\cite{cress2022a9} & surv & 5.4k & 3 & \checkmark & \checkmark & $\times$ \\ 
LUMPI~\cite{LUMPI}        & surv+ego  & 200k  & 3 & \checkmark & \checkmark & $\times$ \\
\midrule
\textbf{Detailed categories} & & & & & & \\
Sonia Dávila-Soberón et al.~\cite{davila2025novel}  & surv  & 2.8k  & 3 & \checkmark & \checkmark & \checkmark \\
Indoor hospital~\cite{kollmitz2019deep} & ego  & 17k   & 5 & $\times$  & \checkmark & \checkmark \\
X-world~\cite{zhang2021x} & surv  & 72k   & 6 & \checkmark & $\times$  & \checkmark \\
PMMA (Ours)  & surv  & 28k   & 9 & \checkmark & \checkmark & \checkmark \\
\bottomrule
\end{tabular}
\begin{tablenotes}
\footnotesize
    \item \hspace{6em} \textbf{Notes:} surv and ego denote surveillance and vehicle–mounted views, respectively.
\end{tablenotes}
\end{table*}  
\end{landscape}
\section{The PMMA Dataset}
\label{sec:PMMA_Dataset}
\subsection{Data Collection}
\subsubsection{Acquisition and Ethics}
The data collection for this dataset was approved by the ethics committee (``Comité d'éthique à la recherche avec des êtres humains'') of Polytechnique Montréal. All participants were volunteers who consented to be filmed and have their data included in the dataset. None of the participants are actual users of mobility aids; instead, they are graduate students performing scenarios to simulate behaviors with mobility aids. Volunteers change and use different mobility aids within each session. 

\subsubsection{Location} 
\begin{figure}[!t]
    \centering
    \includegraphics[width=\linewidth]{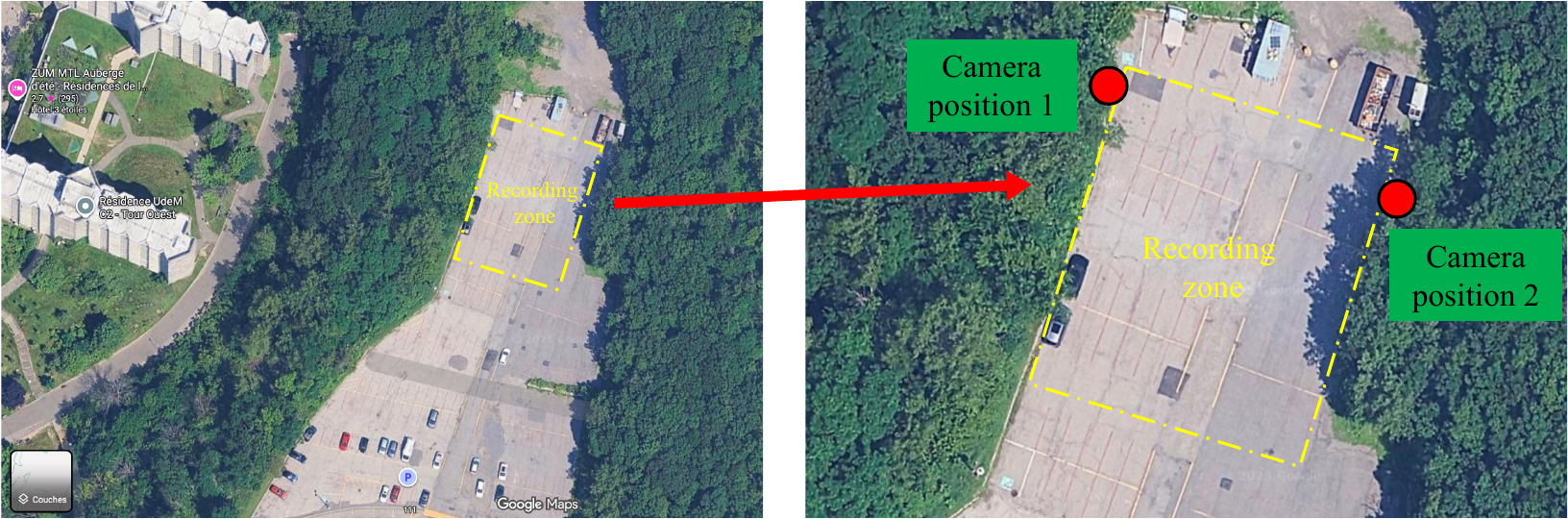}
    \caption{Data collection area in the parking lot of Polytechnique Montréal}
    \label{fig:collection_spot}
\end{figure}

The dataset was recorded from two camera positions in an outdoor parking lot of Polytechnique Montréal, in Montréal, Québec, Canada. The recording zone captured from Google Maps is shown in Figure~\ref{fig:collection_spot}. We focus on the area delineated by traffic cones, which marks the boundary of the region of interest in the scene, highlighted in Figure~\ref{fig:example_of_our_dataset}.

\subsubsection{Camera}
\begin{figure}[!t]
    \centering
    \includegraphics[width=0.5\linewidth]{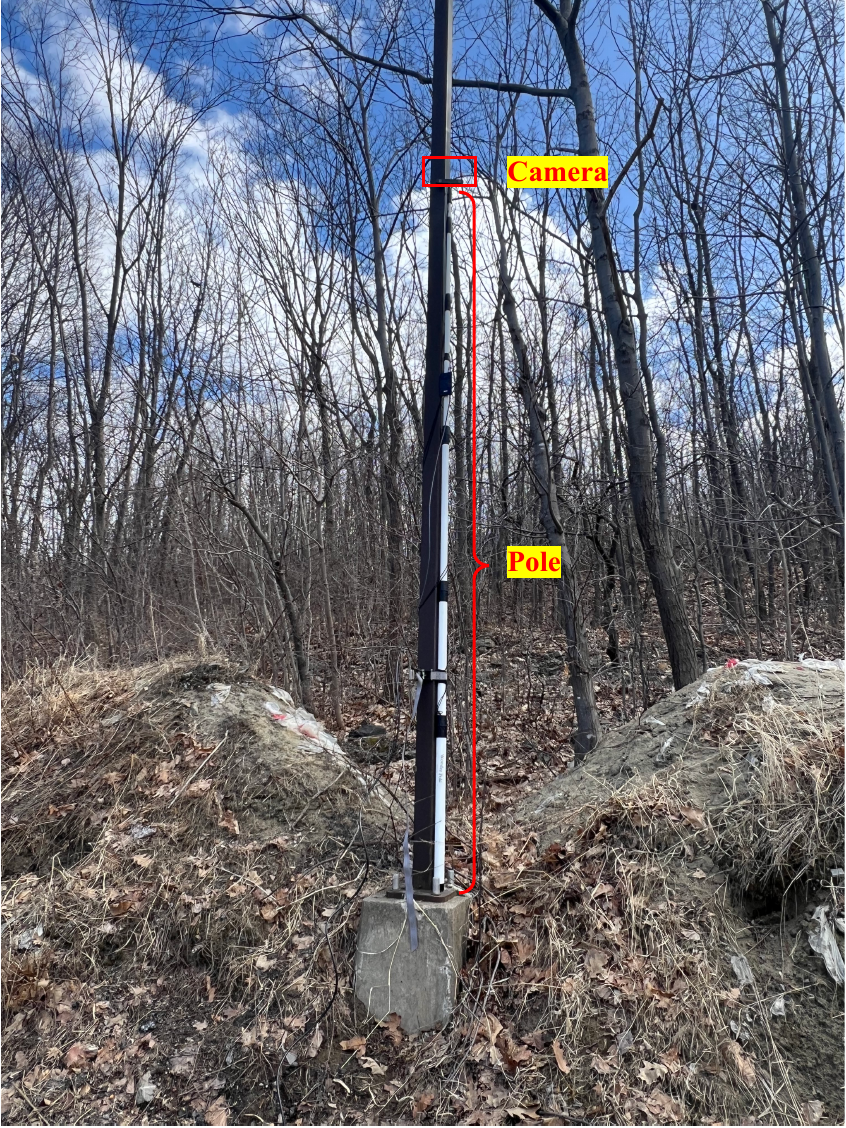}
    \caption{Illustration of the camera and the pole}
    \label{fig:fig1_camera_and_pole}
\end{figure}

The dataset was recorded utilizing a stereo camera ZED 2~\cite{zed2}. It provides dual streams in a side-by-side configuration with resolution $2208 \times 1242$ at 15 frame per second (fps) and has a maximum field of view (FOV) of horizontal $\times$ vertical $\times$ diagonal angles of $110^\circ \times 70^\circ \times 120^\circ$. The camera was mounted on a 4-m-high pole, which was mounted on two distinct streetlight poles giving two separate viewpoints: one of the installations is shown in Figure~\ref{fig:fig1_camera_and_pole}. A Linux system laptop was used to record the video stream from the camera via the VLC media player~\cite{vlc}.

\subsubsection{Recording sessions} Three recording sessions were conducted on the same day. The sunlight varies throughout the recording period due to partially cloudy weather conditions.
The recording sessions are summarized in Table~\ref{tab:details_video_measurements}.

\subsubsection{Object categories} The mobility aids utilized in this data collection includes three wheelchairs, two walkers and two canes. With these mobility aids, the dataset includes nine pedestrian categories, representing a variety of mobility aids and users, shown in Table~\ref{tab:pedestrian_categories}.
Going into details, the walker category is further refined into two subcategories based on the activity status, walking or resting. These subcategories allow for more fine-grained behavior analysis, though they can be easily merged into a single walker category if needed.
For wheelchair-related instances, we introduce a more structured annotation strategy. Specifically, we consider a wheelchair and the associated person(s) as a single unit and classify them into five distinct subcategories: a person alone in a wheelchair, a person pushing an empty wheelchair, a person pushing a person in a wheelchair as a group and its components separately, the person pushing and the seated person.

This hierarchical categorization allows us to model complex pedestrian interactions and mobility aid usage more accurately, while maintaining the flexibility to merge classes as needed for specific tasks or evaluation protocols.



\subsection{Annotation}
\subsubsection{Video pre-processing} Each stereo video is saved to a sequence of stereo images firstly. Each stereo image is then separated into left and right images, and the left images are annotated.

\subsubsection{Annotation tool} We use the computer vision annotation tool (CVAT)~\cite{CVAT_ai_Corporation_Computer_Vision_Annotation_2023}, which is an online platform for annotating object detection and tracking. The annotations are in the COCO format to facilitate model training and evaluation.

\subsubsection{Annotation details} The annotations were separated into three tasks with respect to the three capture sessions. The annotations were initially generated by labelling key frames at every 100 frames using the interpolation-based tracking tool. Linear interpolation was applied between key frames to estimate intermediate object locations, followed by manual frame-by-frame corrections to ensure accuracy.

We also include occlusion attributes in the annotations. Following common occlusion conventions, such as those used in KITTI~\cite{Geiger2012CVPR}, we assign labels 0, 1, and 2 to represent no occlusion, partial occlusion, and full occlusion, respectively. Additionally, we introduce an extra label, 3, to indicate objects in the tree shadows, which is not defined in the original KITTI annotation scheme.

\begin{landscape}
\begin{table}[!t]
    \centering
    \caption{Details of the collected video recordings}
    \label{tab:details_video_measurements}
    \begin{tabular}{ccccc}
        \toprule
        \textbf{Recording session} & \textbf{Duration} & \textbf{FPS} & \textbf{Position} & \textbf{Number of Frames} \\
        \midrule
        Video~1 & 10 min & 15 & 1 & 9,000 \\
        Video~2 & 10 min & 15 & 1 & 5,000 \\
        Video~3 & 15 min & 13 & 2 & 14,637 \\
        \bottomrule
    \end{tabular}
\end{table}


\begin{table*}[!t]
\centering
\caption{Description of pedestrian-related categories}
\label{tab:pedestrian_categories}
\begin{tabular}{>{\centering\arraybackslash}m{1.5cm} m{4cm} m{9cm}}
\toprule
\textbf{Icon} & \textbf{Category} & \textbf{Description} \\
\midrule
\includegraphics[width=0.6cm]{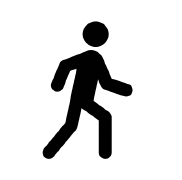} & Ped & Pedestrian without mobility aid \\
\includegraphics[width=0.6cm]{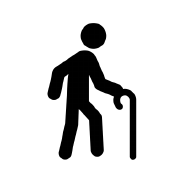} & Cane & Pedestrian using a cane \\
\includegraphics[width=0.6cm]{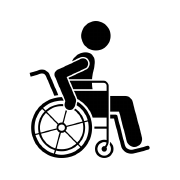} & Wheelchair & Person in a self-propelled wheelchair \\
\includegraphics[width=0.6cm]{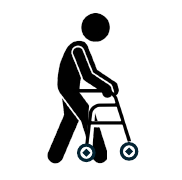} & WalkerWalking & Person walking with a walker \\
\includegraphics[width=0.6cm]{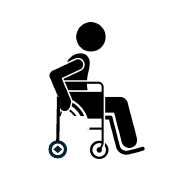} & WalkerResting & Person resting with a walker \\

\includegraphics[width=0.6cm]{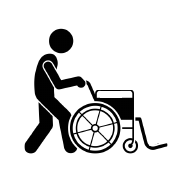} & PushEmptyWheelchair & Person pushing an empty wheelchair \\
\includegraphics[width=0.6cm]{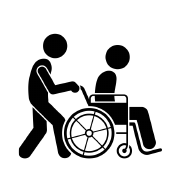} & 
WheelchairGroup
& Person pushing a person in a wheelchair as a group \\
\includegraphics[width=0.6cm]{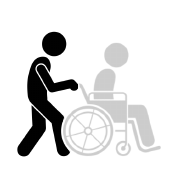} & 
WheelchairPusher
& Person pushing only (from WheelchairGroup) \\
\includegraphics[width=0.6cm]{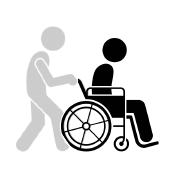} & 
WheelchairPushedUser
& Person in a wheelchair only (from WheelchairGroup) \\
\bottomrule
\end{tabular}
\end{table*}
\end{landscape}
\subsection{Statistics} 
\begin{figure}
    \centering
    \includegraphics[width=\linewidth]{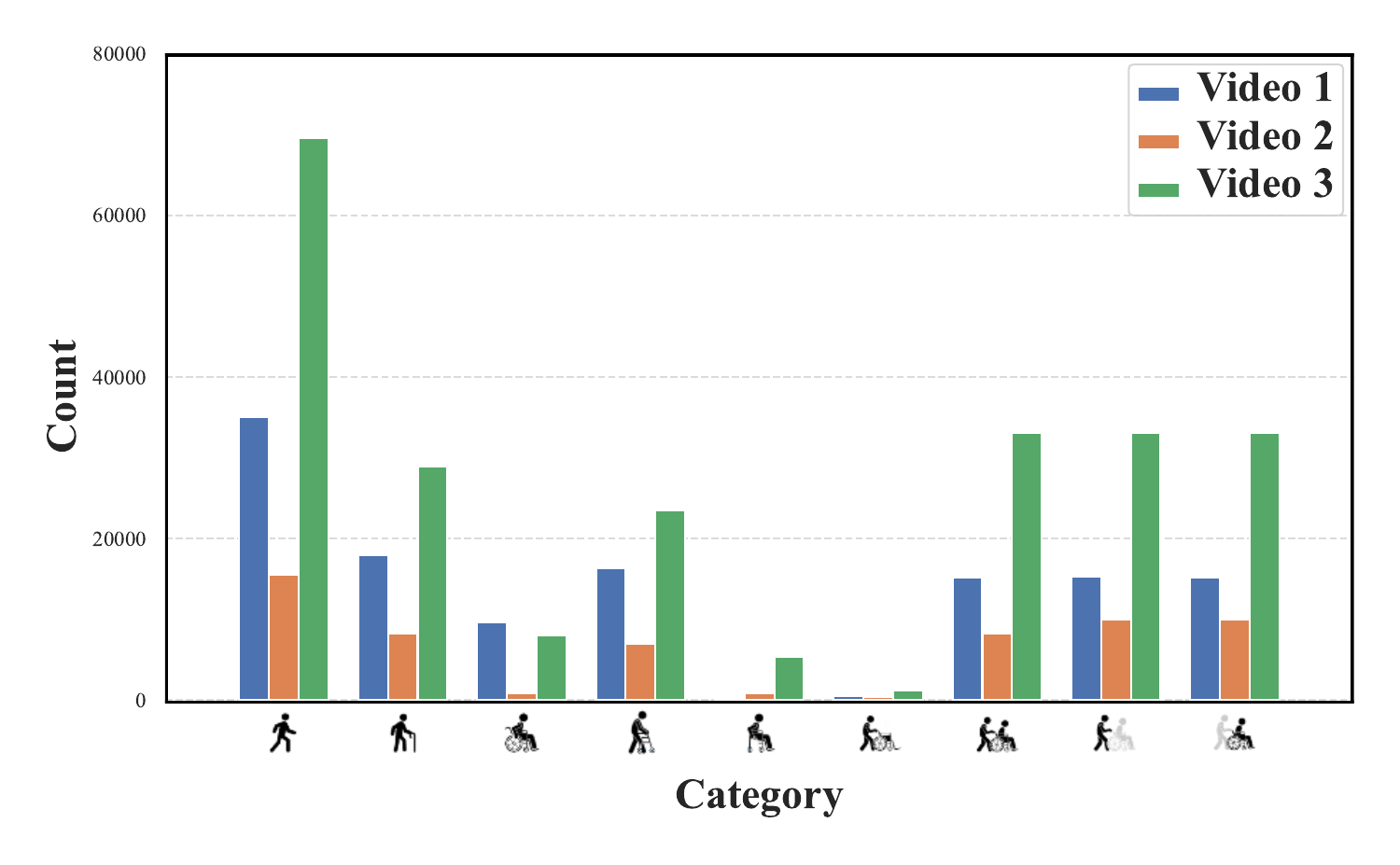}
    \caption{Category-wise annotation counts for each video}
    \label{fig:annotation_statistics}
\end{figure}

Around 30,000 images were annotated in our PMMA dataset. 
Figure~\ref{fig:annotation_statistics} illustrates the annotation counts for each category per video. 
The frame distributions for the training, validation and test sets are shown in Table~\ref{tab:frame_distribution}. For the tracking task, the two video clips in the test set are used to evaluate the tracker performance.

\begin{table}[!t]
    \centering
    \caption{Frame distribution across training, validation, and test sets from different video sources}
    \label{tab:frame_distribution}
        \begin{tabular}{l@{\hskip 4pt}c@{\hskip 4pt}c@{\hskip 8pt}c@{\hskip 4pt}c@{\hskip 8pt}c@{\hskip 4pt}c}
        \toprule
         & \multicolumn{2}{c}{\textbf{Train}} & \multicolumn{2}{c}{\textbf{Validation}} & \multicolumn{2}{c}{\textbf{Test}} \\
        \cmidrule(lr){2-3} \cmidrule(lr){4-5} \cmidrule(lr){6-7}
        \textbf{Video} & Video 1 & Video 3 & Video 2 & Video 3 & Video 2 & Video 3 \\
        \midrule
        \textbf{Number} & 9,000 & 13,248 & 1,700 & 534 & 4,357 & 799 \\
        \textbf{Total}  & \multicolumn{2}{c}{22,248} & \multicolumn{2}{c}{2,234} & \multicolumn{2}{c}{5,156} \\
        \bottomrule
        \end{tabular}
\end{table} 

\begin{table}[!t]
\centering
\caption{Comparison of object detection models}
\label{tab:model_comparison}
\resizebox{\textwidth}{!}{  
\begin{tabular}{lcccc}
\toprule
\textbf{Model}  & \textbf{Type} & \textbf{Stage} & \textbf{Anchor-free} & \textbf{Backbone} \\
\midrule
Faster R-CNN~\cite{faster-rcnn} & CNN  & Two-stage & No  & ResNet-50 \\
CenterNet~\cite{centernet} & CNN & One-stage & Yes & Hourglass \\
YOLOX~\cite{yolox}  & CNN & One-stage & Yes & CSPDarknet \\
DETR~\cite{detr} & Transformer & One-stage & Yes & ResNet-50 \\
Deformable DETR~\cite{deformable-detr}  & Transformer & One-stage & Yes & ResNet-50 \\
DINO~\cite{dino}  & Transformer & One-stage & Yes & ResNet-50 \\
RT-DETR~\cite{rt_detr_v2}  & Transformer & One-stage & Yes & ResNet-50 \\
\bottomrule
\end{tabular}
}
\end{table}

\section{Experimental Methodology}
\label{sec:Experiments}
\subsection{Introduction}

To establish baseline performance on our dataset, we conducted a series of experiments comparing the performance of object detection and MOT methods. For object detection, we used the MMDetection~\cite{MMDetection} framework and selected seven representative methods: Faster R-CNN~\cite{faster-rcnn}, CenterNet~\cite{centernet}, YOLOX~\cite{yolox}, DETR~\cite{detr}, Deformable DETR~\cite{deformable-detr}, DINO~\cite{dino} and RT-DETR~\cite{rt_detr_v2}, as summarized in Table~\ref{tab:model_comparison}. All models were pretrained on the COCO dataset and then trained on our dataset to evaluate their detection performance across the nine pedestrian-related categories.

For MOT, we applied three trackers, including ByteTrack~\cite{bytetrack}, BOT-SORT~\cite{bot_sort} and OC-SORT~\cite{oc_sort}, on the detection results, as summarized in Table~\ref{tab:mot_comparison}. This experimental setup allows us to establish baseline detection and tracking performance on the dataset and provides a reference for future research using our data.

For object detection, we trained for 50 epochs and we use early-stopping. We used an AdamW~\cite{loshchilov2019adamw} optimizer with a learning rate of $1\times 10^{-4}$ and a weight decay of $1\times 10^{-4}$. 

\begin{figure*}
    \centering
    \includegraphics[width=\linewidth]{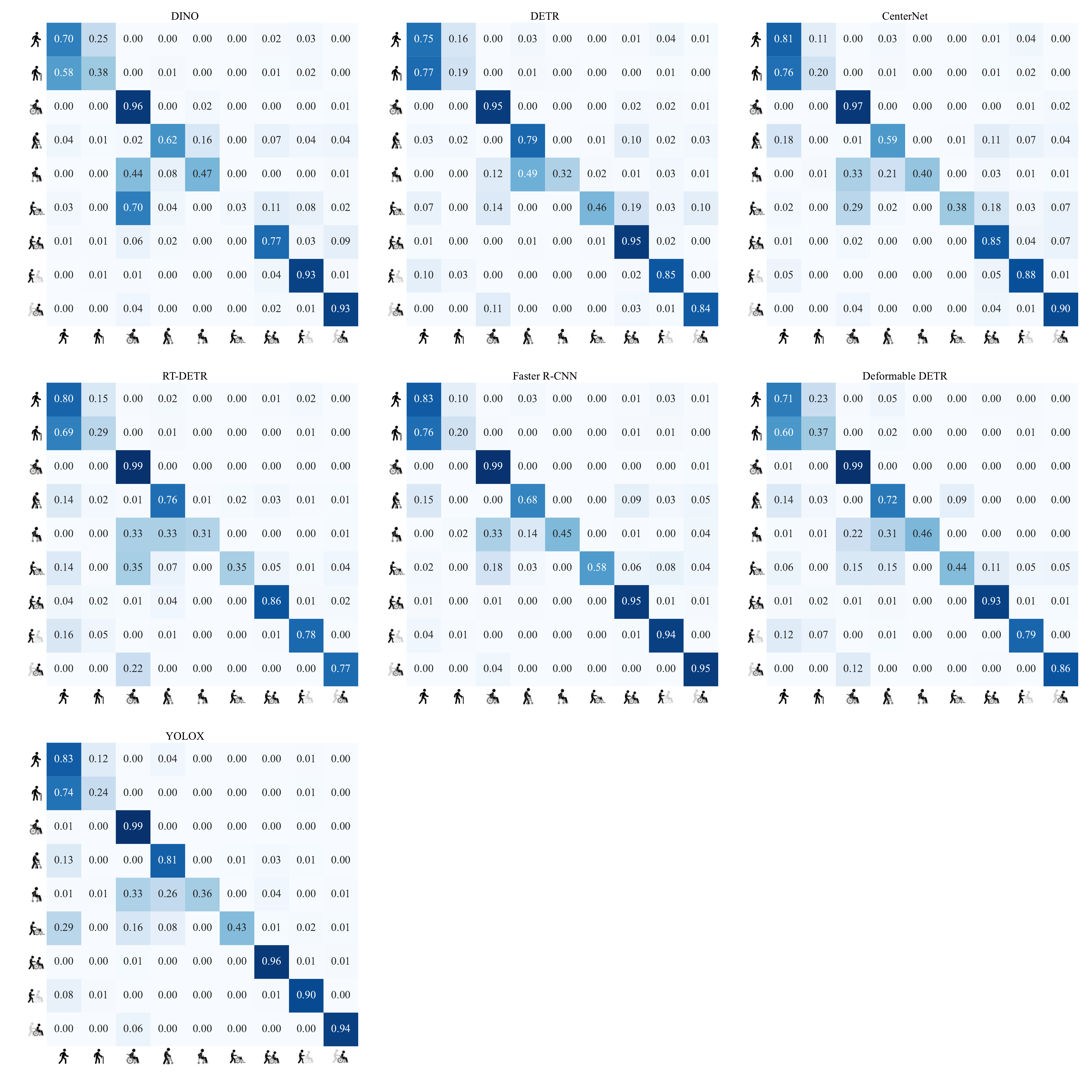}
    \caption{Row-normalized confusion matrices of all detection methods on the test set. Each row and column represent the ground truth (GT) and predicted classes, respectively}
    \label{fig:confusion_matrix}
\end{figure*}

\subsection{Evaluation Metrics}
\subsubsection{Object Detection}
We evaluated the performance of the detection methods using the standard COCO metrics~\cite{coco_metrics}:
\begin{itemize}
    \item \textbf{mAP:} The mean of average precision (AP), where AP is computed by integrating the precision--recall curve at a fixed IoU threshold, measuring overall detection performance.
  \item \textbf{AP$_{50}$:} Average precision over multiple recall levels at an Intersection over Union (IoU) threshold of 0.5, measuring detection accuracy when predicted boxes overlap with ground truth by at least 50\%.
  \item \textbf{AP$_{75}$:} Average precision over multiple recall levels at an IoU threshold of 0.75, reflecting stricter overlap criteria for more precise detections.
  \item $\textbf{AP}_{S}$, $\textbf{AP}_{M}$, $\textbf{AP}_{L}$: Average precision scores over multiple recall levels for small (area $\leq$ 32$^2$ pixels), medium (32$^2$ $<$ area $<$ 96$^2$ pixels), and large (area $\ge$ 96$^2$ pixels) objects, respectively, assessing performance across different object scales.
  \item \textbf{mAR:} The mean average recall over multiple IoU thresholds, indicating the model's ability to find all relevant objects.
  \item $\textbf{AR}_{S}$, $\textbf{AR}_{M}$, $\textbf{AR}_{L}$: Average recall for small, medium, and large objects, respectively, where the object size definitions are the same as those used for $\text{AP}_{S}$, $\text{AP}_{M}$, and $\text{AP}_{L}$.
  \item \textbf{mAP} and \textbf{mAR} are also reported across different occlusion levels, including no occlusion, partial occlusion, and full occlusion. Objects in shadows are included in the Full occlusion category.
\end{itemize}

All evaluation metrics are reported for the overall performance across the nine categories, whereas for the category-wise analysis, only mAP, AP$_{50}$, and AP$_{75}$ are presented.

\subsubsection{Object Tracking} We adopted the Higher Order Tracking Accuracy (HOTA) metric~\cite{HOTA} for each test set. HOTA is a comprehensive metric that jointly considers detection accuracy (DetA), association accuracy (AssA), and localization accuracy (LocA).
HOTA is defined as follows:
\begin{equation}
    \text{HOTA}_\alpha = \sqrt{ \frac{ \sum_{c \in \{\text{TP}\}} \mathcal{A}(c) }{ |\text{TP}| + |\text{FN}| + |\text{FP}| } }
\end{equation}
where $|\text{TP}|$, $|\text{FN}|$, and $|\text{FP}|$ denote the number of true positives, false negatives, and false positives, respectively.  
Each $c$ represents a true positive match between a ground truth ID and a predicted ID.  
$\mathcal{A}(c)$ denotes the \emph{association accuracy} of match $c$, which is defined as:

\begin{equation}
\mathcal{A}(c) = \frac{ |\text{TPA}(c)| }{ |\text{TPA}(c)| + |\text{FNA}(c)| + |\text{FPA}(c)| }
\end{equation}
where $|\text{TPA}(c)|$, $|\text{FNA}(c)|$, and $|\text{FPA}(c)|$ refer to the number of true positive, false negative, and false positive associations related to match $c$, respectively.
\begin{landscape}
\begin{table}[h]
\centering
\caption{Comparison of tracking methods (Re-ID refers to the use of appearance-based features to associate detections across frames, enabling identity recovery after occlusion or long-term target loss.)}
\label{tab:mot_comparison}
\begin{tabular}{lcccc}
\toprule
\textbf{Model} & \textbf{Re-ID} & \textbf{Detection-based Tracking}  \\
\midrule
ByteTrack~\cite{bytetrack} & No  & Yes  \\
BOT-SORT~\cite{bot_sort}  & Yes & Yes \\
OC-SORT~\cite{oc_sort}   & No  & Yes \\
\bottomrule
\end{tabular}
\end{table}

\begin{table}[h]
\centering
\tiny
\caption{Detection performance evaluated on the test set across all categories}
\label{tab:object_performance_all_test}
\begin{tabular}{lcccccccccc}
\toprule
\textbf{Method}

& \textbf{mAP($\%$)} & \textbf{AP$_{50}$($\%$)} & \textbf{AP$_{75}$($\%$)} & \textbf{AP$_S$($\%$)} & \textbf{AP$_M$($\%$)} & \textbf{AP$_L$($\%$)} & \textbf{mAR($\%$)} & \textbf{AR$_S$($\%$)} & \textbf{AR$_M$($\%$)} & \textbf{AR$_L$($\%$)} \\
\midrule
DINO & 27.8 & 55.6 & 24.1 & 7.2 & 24.9 & 42.2 & 57.5 & 9.9 & 53.2 & \textbf{73.1} \\
DETR & 30.4 & 60.7 & 26.7 & 0.0 & 29.7 & 38.6 & 45.4 & 0.3 & 44.8 & 52.7 \\
RT-DETR & 40.0 & 59.3 & 48.6 & 19.3 & 38.1 & 43.0 & 47.3 & 21.3 & 44.8 & 51.1 \\
CenterNet & 44.4 & 67.7 & 52.5 & \textbf{46.9} & 48.9 & \underline{45.0}  & \underline{60.7}  & \textit{47.0}  & \textit{62.6} & 61.4 \\
Deformable DETR & \underline{47.4}  & \underline{68.4}  & \underline{58.4} & \textit{22.2}  & \textit{50.4} & 42.6 & \textbf{64.7}  & \textbf{54.4}  & \textbf{65.5}  & \textit{64.5}  \\
Faster R-CNN & \textit{49.4}  & \textbf{73.9}  & \textit{60.5}  & \underline{19.8}  & \textbf{51.5}  & \textbf{49.5}  & \underline{58.8}  & 21.9 & \underline{59.8}  & \underline{59.5} \\
YOLOX & \textbf{49.5}  & \textit{69.5} & \textbf{61.9}  & 14.8 & \underline{49.9}  & \textit{49.0}  & 57.4 & \underline{32.8}  & 58.8 & 56.0 \\
\bottomrule
\end{tabular}
\begin{tablenotes}
\footnotesize
    \item \textbf{Notes:} \textbf{Bold}, \textit{italic}, and \underline{underline} represent the first, second, and third best performances, respectively. Methods are sorted based on mAP. These notational conventions apply to all the following tables in this paper.
\end{tablenotes}
\end{table} 
\begin{table}[h]
\centering
\tiny
\caption{Detection performance for each of the nine categories on the test set}
\label{tab:object_performance_sep_test}
\scriptsize
\begin{tabular}{c}
\toprule

\begin{tabular}{lccc|ccc|ccc}
\textbf{Method} 
& \multicolumn{3}{c|}{
    \raisebox{-0.20\height}{\includegraphics[width=0.35cm]{icons/1_ped.pdf}}
     Ped
}
& \multicolumn{3}{c|}{
    \raisebox{-0.20\height}{\includegraphics[width=0.35cm]{icons/2_cane.pdf}}
     Cane
}
& \multicolumn{3}{c}{
    \raisebox{-0.20\height}{\includegraphics[width=0.35cm]{icons/5_wheelchair.pdf}}
     Wheelchair
}
\\
& \textbf{mAP($\%$)} & \textbf{AP$_{50}$($\%$)} & \textbf{AP$_{75}$($\%$)} & \textbf{mAP($\%$)} & \textbf{AP$_{50}$($\%$)} & \textbf{AP$_{75}$($\%$)} & \textbf{mAP($\%$)} & \textbf{AP$_{50}$($\%$)} & \textbf{AP$_{75}$($\%$)} \\
\midrule
DETR & 23.1 & 51.7 & 16.4 & 4.7 & 11.2 & 3.1 & 51.1 & 76.8 & 63.6 \\
DINO & 27.8 & 58.1 & 22.6 & 9.0 & \textbf{26.5} & 5.3 & 34.9 & 66.0 & 20.0 \\
RT-DETR & 37.2 & 60.6 & 41.1 & \underline{9.2} & 16.3 & \textit{9.8} & 59.0 & 78.7 & 73.7 \\
Faster R-CNN & 36.5 & 61.3 & 38.9 & 8.7 & 16.6 & 7.8 & \underline{62.2} & \underline{86.4} & \underline{79.4} \\
CenterNet & \underline{38.5} & \textit{63.1} & \underline{42.0} & \textit{10.2} & \underline{18.9} & \underline{9.5} & 46.0 & 62.7 & 61.1 \\
Deformable DETR & \textit{39.7} & \underline{62.3} & \textit{47.3} & \textbf{14.0} & \textit{26.0} & \textbf{13.9} & \textit{69.5} & \textit{87.7} & \textit{86.9} \\
YOLOX & \textbf{44.1} & \textbf{66.2} & \textbf{53.7} & 8.8 & 16.2 & 9.0 & \textbf{72.2} & \textbf{88.9} & \textbf{88.4} \\
\end{tabular} \\

\begin{tabular}{lccc|ccc|ccc}
\toprule
\textbf{Method} 
& \multicolumn{3}{c|}{
    \raisebox{-0.20\height}{\includegraphics[width=0.35cm]{icons/3_walkerWalking.pdf}}
     WalkerWalking
}
& \multicolumn{3}{c|}{
    \raisebox{-0.20\height}{\includegraphics[width=0.35cm]{icons/4_walkerResting.pdf}}
     WalkerResting
}
& \multicolumn{3}{c}{
    \raisebox{-0.20\height}{\includegraphics[width=0.35cm]{icons/6_pushEmptyWheelchair.pdf}}
     PushEmptyWheelchair
}
\\
& \textbf{mAP($\%$)} & \textbf{AP$_{50}$($\%$)} & \textbf{AP$_{75}$($\%$)} & \textbf{mAP($\%$)} & \textbf{AP$_{50}$($\%$)} & \textbf{AP$_{75}$($\%$)} & \textbf{mAP($\%$)} & \textbf{AP$_{50}$($\%$)} & \textbf{AP$_{75}$($\%$)} \\
\midrule
DINO & 8.4 & 12.4 & 11.1 & 39.1 & 68.6 & 42.4 & 12.7 & 38.0 & 2.3 \\
DETR & 24.1 & 43.7 & 23.7 & 37.6 & 71.0 & 38.1 & 24.3 & 44.7 & 27.8 \\
RT-DETR & 25.7 & 35.8 & 34.0 & \underline{48.0} & \underline{73.3} & 56.1 & 21.6 & 35.2 & 25.8 \\
CenterNet & \textit{41.4} & \textit{55.0} & \textit{54.2} & 47.9 & 70.7 & \underline{56.4} & \underline{35.4} & \textit{60.9} & 38.8 \\
Deformable DETR & 23.8 & 33.1 & 30.3 & 45.6 & 73.1 & 50.9 & \textit{40.5} & \underline{55.8} & \textit{52.4} \\
YOLOX & \underline{34.4} & \underline{46.9} & \underline{44.3} & \textbf{53.9} & \textbf{79.5} & \textbf{62.6} & 27.3 & 39.5 & \underline{38.9} \\
Faster R-CNN & \textbf{52.2} & \textbf{71.6} & \textbf{70.3} & \textit{48.6} & \textit{77.7} & \textit{56.9} & \textbf{44.2} & \textbf{61.3} & \textbf{58.6} \\
\end{tabular} \\
\toprule

\begin{tabular}{lccc|ccc|ccc}
\textbf{Method} 
& \multicolumn{3}{c|}{
    \raisebox{-0.20\height}{\includegraphics[width=0.35cm]{icons/7_WheelchairGroup.pdf}}
     WheelchairGroup
}
& \multicolumn{3}{c|}{
    \raisebox{-0.20\height}{\includegraphics[width=0.35cm]{icons/8_WheelchairPusher.pdf}}
     WheelchairPusher
}
& \multicolumn{3}{c}{
    \raisebox{-0.20\height}{\includegraphics[width=0.35cm]{icons/9_WheelchairPushedUser.pdf}}
     WheelchairPushedUser
}
\\
& \textbf{mAP($\%$)} & \textbf{AP$_{50}$($\%$)} & \textbf{AP$_{75}$($\%$)} & \textbf{mAP($\%$)} & \textbf{AP$_{50}$($\%$)} & \textbf{AP$_{75}$($\%$)} & \textbf{mAP($\%$)} & \textbf{AP$_{50}$($\%$)} & \textbf{AP$_{75}$($\%$)} \\
\midrule
DETR & 42.2 & 91.5 & 25.4 & 26.3 & 73.5 & 11.3 & 40.7 & 82.4 & 30.8 \\
DINO & 51.9 & 80.4 & 64.8 & 30.7 & 68.9 & 23.9 & 35.9 & 81.6 & 24.0 \\
RT-DETR & 60.8 & 79.9 & 77.4 & 47.3 & 75.5 & 55.6 & 50.7 & 78.1 & 63.9 \\
CenterNet & 68.7 & 92.2 & 85.0 & 56.5 & \textit{92.5} & 64.2 & 55.1 & 93.0 & 61.2 \\
Deformable DETR & \underline{72.0} & \underline{94.8} & \underline{91.7} & \underline{57.6} & 89.8 & \textit{69.6} & \textit{64.3} & \underline{93.3} & \textit{82.3} \\
Faster R-CNN & \textit{74.0} & \textit{97.1} & \textit{93.3} & \textit{58.5} & \textbf{95.9} & \underline{68.0} & \underline{59.7} & \textit{97.3} & \underline{71.7} \\
YOLOX & \textbf{74.4} & \textbf{98.3} & \textbf{95.0} & \textbf{59.9} & \underline{92.3} & \textbf{74.0} & \textbf{70.8} & \textbf{97.6} & \textbf{90.9} \\
\end{tabular} \\
\bottomrule
\end{tabular}
\end{table}

\end{landscape}
Specifically, we report HOTA along with DetA, AssA, and LocA, presenting both the overall performance across all nine categories and the category-wise performance. We also include the total number of detected objects and tracked identities.


\section{Results and Discussion}
\label{sec:ResultsAndDiscussions}
\subsection{Object Detection}
The object detection performance across all categories is given in Table~\ref{tab:object_performance_all_test}. YOLOX, Deformable DETR, and Faster R-CNN achieve the top three performances across most metrics. Specifically, YOLOX has the best AP while Deformable DETR has the best AR. However, the performance differences among CenterNet, RT-DETR, Faster R-CNN, Deformable DETR, and YOLOX are relatively small, whereas DINO and DETR lag significantly behind the other five.

Moreover, the object detection performance for each category is given in Table~\ref{tab:object_performance_sep_test}. YOLOX, Deformable DETR, and Faster R-CNN achieve the best performance. CenterNet performs comparably well to these three methods. DETR and DINO consistently exhibit the lowest performance.

Regarding the categories, cane users are the most challenging to detect. This could be attributed to the fact that the cane, being a thin object, is difficult to detect and cane users are therefore confused with pedestrians without any mobility aid. Wheelchair-related categories achieve higher mAP and mAR; however, PushEmptyWheelchair shows the lowest accuracy among them, possibly due to its visual similarity to WheelchairGroup.
Figure~\ref{fig:confusion_matrix} shows that the confusion matrices on the test set are similar for all the methods. There is substantial confusion (1) between pedestrians without any mobility aid and cane users, and (2) between wheelchair users and walker users. 

Table~\ref{tab:occlusion_results} presents the detection results with respect to different occlusion levels. DINO has the best mAP in case of no occlusion and poor performance on the partial and full occlusion cases. Generally, Faster R-CNN, Deformable DETR and YOLOX have the top performance consistent with the overall detection results. 

\begin{table}[t!]
\centering
\caption{Detection results on the test set with respect to different occlusion levels}
\label{tab:occlusion_results}
\begin{tabular}{l ccc ccc}
\toprule
\multirow{2}{*}{\centering \textbf{Methods}} 
& \multicolumn{3}{c}{\textbf{mAP}($\%$)} & \multicolumn{3}{c}{\textbf{mAR}($\%$)} \\
\cmidrule(lr){2-4} \cmidrule(lr){5-7}
 & No & Partial & Full & No & Partial & Full \\
\midrule
DETR & 45.20 & 10.34 &  8.33 & 70.56 & 72.63 & 63.38 \\
CenterNet & 50.99 & \underline{12.37} & 10.04 & \underline{76.79} & \underline{84.42} & \underline{75.04} \\
RT-DETR  & 52.50 & 11.28 & \textbf{11.19} & 71.87 & 69.35 & 73.82 \\
Deformable DETR  & 53.83 & 11.91 & \underline{10.89} & 70.32 & 69.36 & 68.25 \\
Faster R-CNN    & \underline{54.10} & \textit{13.03} & 10.79 & \textbf{81.85} & \textbf{88.36} & \textbf{81.02} \\
YOLOX  & \textit{55.98} & \textbf{13.38} & \textit{11.05} & \textit{80.22} & \textit{85.28} & \textit{78.11} \\
DINO & \textbf{57.06} & 10.56 &  5.72 & 57.56 & 48.73 & 29.48 \\
\bottomrule
\end{tabular}
\end{table}

\subsection{Object Tracking}
Based on the detection performance, we applied three tracking algorithms to the top five methods, excluding DINO and DETR. We report the object tracking results on overall performance and per-category performance separately on videos~2 and 3 from the test set in Table~\ref{tab:tracking_performance}.

For overall performance, among the detectors, YOLOX achieves the best HOTA on video 2, while Deformable DETR performs best on video 3. Among the trackers, ByteTrack consistently outperforms the others on both videos, though the differences between the three trackers are relatively small and OC-SORT with Faster R-CNN on video 2 and BOT-SORT with Deformable DETR on video 3 show good performance. 
All trackers show strong DetA and LocA. The AssA for video 2 is relatively low, primarily because the tracker generates more IDs than the ground truth objects, resulting in a reduced overall HOTA score.

The category-wise performances are summarized in Tables~\ref{tab:all_categories} to~\ref{tab:all_categories_video_3_part_3}. On video 2, the performance of all methods on the categories Cane, WalkerResting, and PushEmptyWheelchair is lower than 10~\%; The performance on the categories Ped and WalkerWalking is between 10~\% and 40~\%; the performance on Wheelchair, WheelchairGroup, WheelchairPusher and WheelchairPushedUser is the highest, with the best performance over 40~\%. YOLOX and Deformable DETR have quite close performance. On video 3, all the methods perform better on Ped, Cane, Wheelchair, WheelchairGroup, WheelchairPusher and WheelchairPushedUser, with the best performance over 60~\%. 
Deformable DETR and YOLOX demonstrate the strongest overall performance across most mobility-related categories.
On Video 2, Deformable DETR ranks within the top three for 6 of the 9 categories, while YOLOX achieves top-three performance in all 9 categories.
On Video 3, Deformable DETR attains top-three results in 8 of the 8 evaluated categories, and YOLOX performs similarly with 7 top-three finishes.
Overall, the performance differences among the trackers are relatively small; detection accuracy remains the key factor influencing tracking performance.

\begin{table*}[!t]
\centering
\caption{Tracking performance across different detectors and trackers for all categories on videos 2 and 3 from the test set, sorted by HOTA}
\label{tab:tracking_performance}
\resizebox{\textwidth}{!}{
\begin{tabular}{c}
\toprule
\begin{tabular}{l l c c c c r r r r}
\multicolumn{9}{c}{\textbf{Video 2}}  \\
\toprule
\multirow{2}{*}{\textbf{Detector}} & \multirow{2}{*}{\textbf{Method}} & \multicolumn{4}{c}{\textbf{Tracking Metrics (\%)}} & \multicolumn{2}{c}{\textbf{Detection Counts}} & \multicolumn{2}{c}{\textbf{Track ID Counts} \vspace{1mm}} \\
\cmidrule(lr){3-6} \cmidrule(lr){7-8} \cmidrule(lr){9-10}
&  & \textbf{HOTA} & \textbf{DetA} & \textbf{AssA} & \textbf{LocA} & \textbf{Predictions} & \textbf{GT} & \textbf{Predictions} & \textbf{GT} \\
\midrule
Deformable DETR   & OC-SORT       & 32.41  & 62.63  & 16.92  & 84.82 & 53,970  & \multirow{15}{*}{64,678} & 453   & \multirow{15}{*}{35} \\
CenterNet         & OC-SORT       & 33.38  & 63.16  & 17.80  & 83.69 & 63,446  &                         & 617   &                     \\
CenterNet         & BOT-SORT      & 34.10  & 60.35  & 19.43  & 83.54 & 74,418  &                         & 1,543 &                     \\
RT-DETR           & OC-SORT       & 34.14  & 63.60  & 18.50  & 82.87 & 60,730  &                         & 468   &                     \\
RT-DETR           & BOT-SORT      & 35.62  & 62.82  & 20.36  & 82.6 & 67,962  &                         & 1,075 &                     \\
RT-DETR           & ByteTrack    & 35.87  & 63.74  & 20.36  & 82.83 & 66,914  &                         & 584   &                     \\
CenterNet         & ByteTrack    & 36.27  & 61.29  & 21.61  & 83.75 & 72,990  &                         & 668   &                     \\
Deformable DETR   & BOT-SORT      & 37.88  & 67.68  & 21.36  & 84.43 & 64,415  &                         & 1,125 &                     \\
Deformable DETR   & ByteTrack    & 39.00  & 68.03  & 22.55  & 84.7  & 63,348  &                         & 499   &                     \\
Faster R-CNN      & BOT-SORT      & 44.54  & 60.85  & 32.86  & 83.27 & 76,882  &                         & 1,053 &                     \\
Faster R-CNN      & ByteTrack    & 45.34  & 61.76  & 33.59  & 83.27 & 75,869  &                         & 527   &                     \\
Faster R-CNN      & OC-SORT       & 46.39  & 63.70  & 34.12  & 83.39 & 70,059  &                         & 466   &                     \\
YOLOX & BOT-SORT & \underline{52.62} & \underline{67.93} & \textit{41.02} & \underline{85.02} & 71,718 & & 910 & \\
YOLOX & OC-SORT & \textit{53.02} & \textbf{70.07} & \underline{40.38} & \textit{85.14} & 65,796 & & 407 & \\
YOLOX & ByteTrack & \textbf{54.91} & \textit{68.59} & \textbf{44.23} & \textbf{85.21} & 70,864 & & 451 & \\
\bottomrule
\end{tabular}\\[11.5em]

\begin{tabular}{l l c c c c r r r r}
\multicolumn{9}{c}{\textbf{Video 3}} \\
\toprule
\multirow{2}{*}{\textbf{Detector}} & \multirow{2}{*}{\textbf{Method}} & \multicolumn{4}{c}{\textbf{Tracking Metrics (\%)}} & \multicolumn{2}{c}{\textbf{Detection Counts}} & \multicolumn{2}{c}{\textbf{Track ID Counts} \vspace{1mm}} \\
\cmidrule(lr){3-6} \cmidrule(lr){7-8} \cmidrule(lr){9-10}
&  & \textbf{HOTA} & \textbf{DetA} & \textbf{AssA} & \textbf{LocA} & \textbf{Predictions} & \textbf{GT} & \textbf{Predictions} & \textbf{GT} \\
\midrule
CenterNet & OC-SORT & 55.41 & 69.85 & 44.13 & \textbf{94.94} & 13,634  &  \multirow{15}{*}{12,784} & 101  &  \multirow{15}{*}{17} \\
CenterNet & BOT-SORT & 62.50 & 65.34  & 59.97  & 84.89 & 15,374  &  & 290  & \\
CenterNet & ByteTrack & 65.70  & 66.19  & 65.47  & 84.7  & 15,129 & & 108  &  \\
Faster R-CNN  & OC-SORT & 66.89  & 74.43  & 60.38  & 86.02 & 13,490  & & 71 &\\
RT-DETR & OC-SORT  & 69.08  & 76.16  & 63.12  & 85.54 & 12,941  & & 47  &  \\
Faster R-CNN & BOT-SORT & 70.88 & 72.66 & 69.43 & 86.06 & 14,232 & & 187 & \\
Faster R-CNN & ByteTrack & 72.45 & 73.37 & 71.80 & 86.05 & 14,072 & & 94 & \\
YOLOX & ByteTrack & 72.97 & 77.20 & 69.32 & 85.89 & 13,142 & & 84 & \\
YOLOX & BOT-SORT & 73.30 & 76.54 & 70.58 & 85.93 & 13,284 & & 168 & \\
Deformable DETR & OC-SORT & 73.48 & \underline{79.73} & 68.00 & \textit{87.06} & 12,373 & & 36 & \\
YOLOX & OC-SORT & 74.08 & 78.07 & 70.68 & 85.95 & 12,756 & & 70 & \\
RT-DETR & BOT-SORT & 77.56 & 75.82 & \textbf{79.85} & 85.47 & 13,431 & & 115 & \\
RT-DETR & ByteTrack & \underline{77.71} & 76.23 & \textit{79.70} & 85.43 & 13,333 & & 52 & \\
Deformable DETR & BOT-SORT & \textit{79.30} & \textit{80.50} & 78.38 & \underline{87} & 12,747 & & 71 & \\
Deformable DETR & ByteTrack & \textbf{79.42} & \textbf{80.53} & \underline{78.57} & 86.98 & 12,691 & & 34 & \\
\end{tabular}\\
\bottomrule
\end{tabular}
}
\begin{tablenotes}
\footnotesize
    \item \textbf{Note:} Methods are sorted based on HOTA in this table and the following ones. 
\end{tablenotes}
\end{table*}

\section{Conclusion}
\label{sec:Conclusion}

In this study, we collected a mobility aids dataset, named PMMA, and used it to benchmark SOTA detection and tracking methods. PMMA contains high-resolution 
videos with annotations
for nine mobility-aid-related categories, including pedestrians, cane users, two types of walker users (walking and resting), five types of wheelchair users, including wheelchair users, people pushing empty wheelchairs, and three types from wheelchair groups (Wheelchair group, wheelchair pusher and wheelchair user together with wheelchair).
The dataset was divided into training, validation, and test sets. 
We evaluated seven object detectors, including Faster R-CNN, CenterNet, YOLOX, DETR, Deformable DETR, DINO, and RT-DETR, and applied three MOT trackers, namely ByteTrack, BOT-SORT, and OC-SORT, on the top five detection models.
We reported both overall and category-wise performance. The results show that YOLOX, Deformable DETR, and Faster R-CNN achieved the best detection performance, while the differences among the three trackers were small. In terms of categories, Wheelchair, WheelchairGroup, WheelchairPusher, and WheelchairPushedUser were the easiest to detect and track, whereas performance for the remaining categories was less stable and strongly influenced by the specific scenarios.

Overall, YOLOX, Deformable DETR, and Faster R-CNN achieved the best performance among the seven evaluated methods. They performed well on wheelchair-related categories. However, performance on the other categories varies significantly across different scenarios. The differences among the three trackers were minimal, as their results largely depended on the quality of the detections. Nevertheless, no single method performed well across all nine categories, highlighting the ongoing challenge of automatic data collection of mobility aid users. Transportation studies focusing on these users still require improving detection and tracking accuracy.

\section*{Acknowledgments}
The authors gratefully acknowledge financial support from the China Scholarship Council, as well as the financial support of NSERC through the Discovery Grant 2023-05626. The authors would also like to thank the Interuniversity Research Centre on Enterprise Networks, Logistics and Transportation (CIRRELT) for the computing servers and to NVIDIA Corporation for providing GPUs. This research was also enabled in part by support the computing resources provided by Digital Research Alliance of Canada. 

\bibliographystyle{IEEEtran}  
\bibliography{references}  

\begin{table*}[!t]
\centering
\caption{Tracking performance across different detectors and trackers for selected categories on video 2 test set (Part 1)}
\label{tab:all_categories}
\resizebox{\textwidth}{!}{%
\begin{tabular}{c}
\toprule

\begin{tabular}{l l c c c c r r r r}
\multicolumn{9}{c}{
    \raisebox{-0.15\height}{\includegraphics[width=0.4cm]{icons/1_ped.pdf}}
    Ped
} \\
\toprule
\multirow{2}{*}{\textbf{Detector}} & \multirow{2}{*}{\textbf{Method}} & \multicolumn{4}{c}{\textbf{Tracking Metrics (\%)}} & \multicolumn{2}{c}{\textbf{Detection Counts}} & \multicolumn{2}{c}{\textbf{Track ID Counts} \vspace{1mm}} \\ \cmidrule(lr){3-6} \cmidrule(lr){7-8} \cmidrule(lr){9-10} &  & \textbf{HOTA} & \textbf{DetA} & \textbf{AssA} & \textbf{LocA} & \textbf{Predictions} & \textbf{GT} & \textbf{Predictions} & \textbf{GT} \\
\midrule
Deformable DETR & OC-SORT & 17.06 & 26.90 & 10.86 & 83.31 & 16,829 & \multirow{15}{*}{17,383} & 254 & \multirow{15}{*}{9} \\
CenterNet & ByteTrack & 17.67 & 28.57 & 10.97 & 81.76 & 32,880 & & 587 & \\
CenterNet & BOT-SORT & 17.75 & 27.93 & 11.33 & 81.73 & 34,088 & & 1,239 & \\
Faster R-CNN & OC-SORT & 18.27 & 31.22 & 10.74 & 81.54 & 29,371 & & 419 &  \\
Deformable DETR & BOT-SORT & 18.32 & 28.36 & 11.89 & 82.77 & 22,584 & & 713 & \\
Deformable DETR & ByteTrack & 18.40 & 28.86 & 11.81 & 82.98 & 21,903 & & 392 & \\
RT-DETR & OC-SORT & 18.66 & 31.60 & 11.07 & 81.15 & 23,898 & & 342 & \\
Faster R-CNN & ByteTrack & 18.98 & 29.71 & 12.17 & 81.41 & 34,389 & & 482 &\\
Faster R-CNN  & BOT-SORT & 19.40 & 29.12 & 12.98 & 81.36 & 35,277 & & 909 &\\
CenterNet & OC-SORT & 19.73 & 31.64 & 12.35 & 82.08 & 25,376 & & 496 & \\
RT-DETR & BOT-SORT & 20.71 & 29.81 & 14.43 & 80.61 & 29,154 & & 835 & \\
YOLOX  & OC-SORT  & 20.97 & \textbf{34.93} & 12.61 & \textbf{84.15} & 26,140 & & 327 & \\
RT-DETR  & ByteTrack & \underline{21.45} & 30.49 & \underline{15.13} & 80.74 & 28,332 & & 485 & \\
YOLOX & BOT-SORT  & \textit{23.24} & \underline{33.40} & \textbf{16.18} & \underline{83.98} & 31,045 &  & 768 & \\
YOLOX  & ByteTrack & \textbf{23.34} & \textit{33.89} & \textit{16.09} & \textit{84.07} & 30,304 & & 401 & \\
\bottomrule
\end{tabular}\\
\begin{tabular}{l l c c c c r r r r}
\multicolumn{9}{c}{
    \raisebox{-0.15\height}{\includegraphics[width=0.4cm]{icons/2_cane.pdf}}
    Cane
} \\
\toprule
\multirow{2}{*}{\textbf{Detector}} & \multirow{2}{*}{\textbf{Method}} & \multicolumn{4}{c}{\textbf{Tracking Metrics (\%)}} & \multicolumn{2}{c}{\textbf{Detection Counts}} & \multicolumn{2}{c}{\textbf{Track ID Counts} \vspace{1mm}} \\ \cmidrule(lr){3-6} \cmidrule(lr){7-8} \cmidrule(lr){9-10} &  & \textbf{HOTA} & \textbf{DetA} & \textbf{AssA} & \textbf{LocA} & \textbf{Predictions} & \textbf{GT} & \textbf{Predictions} & \textbf{GT} \\
\midrule
Faster R-CNN & OC-SORT & 2.56 & 2.39 & 2.82 & 73.87 & 12,205 & \multirow{15}{*}{7,770} & 407 & \multirow{15}{*}{3} \\
Faster R-CNN & BOT-SORT & 2.71 & 3.17 & 2.42 & 73.47 & 17,642 & & 896 & \\
Faster R-CNN & ByteTrack & 2.75 & 2.98 & 2.65 & 73.88 & 16,717 & & 477 & \\
CenterNet & BOT-SORT & 3.27 & 4.07 & 2.79 & 75.01 & 17,156 & & 1,207 & \\
CenterNet & ByteTrack & 3.34 & 4.08 & 2.89 & 75.37 & 15,812 & & 570 & \\
CenterNet & OC-SORT & 3.55 & 3.85 & 3.40 & 75.67 & 9,499 & & 498 & \\
YOLOX & BOT-SORT & 5.09 & 4.32 & 6.10 & 78.57 & 12,966 & & 758 & \\
YOLOX & ByteTrack & 5.16 & 4.26 & 6.34 & \underline{78.86} & 12,147 & & 399 & \\
Deformable DETR & OC-SORT & 5.25 & \underline{7.47} & 3.72 & \textit{79.25} & 7,783 & & 286 & \\
RT-DETR & ByteTrack & 5.40 & 5.77 & 5.25 & 77.39 & 12,648 & & 463 & \\
YOLOX & OC-SORT & 5.52 & 4.02 & 7.63 & \textbf{79.78} & 8,536 & & 333 & \\
RT-DETR & OC-SORT & 6.01 & 5.54 & \textit{6.85} & 77.59 & 8,785 & & 369 & \\
RT-DETR & BOT-SORT & \underline{6.90} & 5.75 & \textbf{8.94} & 76.91 & 13,583 & & 822 & \\
Deformable DETR & BOT-SORT & \textit{7.18} & \textit{8.32} & 6.21 & 78.45 & 12,964 & & 820 & \\
Deformable DETR & ByteTrack & \textbf{7.31} & \textbf{8.33} & \underline{6.45} & 78.43 & 12,196 & & 407 & \\
\end{tabular}\\
\bottomrule
\begin{tabular}{l l c c c c r r r r}
\multicolumn{9}{c}{
    \raisebox{-0.15\height}{\includegraphics[width=0.4cm]{icons/5_wheelchair.pdf}}
    Wheelchair
} \\
\toprule
\multirow{2}{*}{\textbf{Detector}} & \multirow{2}{*}{\textbf{Method}} & \multicolumn{4}{c}{\textbf{Tracking Metrics (\%)}} & \multicolumn{2}{c}{\textbf{Detection Counts}} & \multicolumn{2}{c}{\textbf{Track ID Counts} \vspace{1mm}} \\ \cmidrule(lr){3-6} \cmidrule(lr){7-8} \cmidrule(lr){9-10} &  & \textbf{HOTA} & \textbf{DetA} & \textbf{AssA} & \textbf{LocA} & \textbf{Predictions} & \textbf{GT} & \textbf{Predictions} & \textbf{GT} \\
\midrule
CenterNet & BOT-SORT & 17.80 & 7.55 & 41.96 & 87.75 & 17,765 & \multirow{15}{*}{1,569} & 1,160 & \multirow{15}{*}{1} \\
CenterNet & ByteTrack & 18.52 & 8.13 & 42.20 & 87.84 & 16,498 & & 557 & \\
RT-DETR & OC-SORT & 23.22 & 13.47 & 40.12 & 85.00 & 9,699 & & 314 & \\
CenterNet & OC-SORT & 23.34 & 13.01 & 41.89 & 87.98 & 10,139 & & 471 & \\
Faster R-CNN & BOT-SORT & 24.46 & 7.54 & 79.51 & 87.22 & 18,034 & & 868 & \\
Faster R-CNN & ByteTrack & 25.16 & 7.94 & 79.79 & 87.47 & 17,172 & & 468 & \\
RT-DETR & BOT-SORT & 25.97 & 9.41 & 71.74 & 84.92 & 13,950 & & 768 & \\
RT-DETR & ByteTrack & 26.75 & 9.97 & 71.84 & 85.12 & 13,170 & & 442 & \\
Faster R-CNN & OC-SORT & 28.86 & 10.49 & 79.53 & 87.26 & 12,889 & & 390 & \\
YOLOX & BOT-SORT & 30.68 & 10.88 & 86.56 & 89.64 & 12,921 & & 730 & \\
YOLOX & ByteTrack & 31.96 & 11.72 & \textbf{87.27} & \textbf{90.04} & 12,081 & & 392 & \\
Deformable DETR & BOT-SORT & 35.48 & 15.08 & 83.55 & 88.86 & 9,167 & & 640 & \\
Deformable DETR & ByteTrack & \underline{36.91} & \textit{16.23} & \underline{84.03} & \underline{ 89.13} & 8,539 & & 360 & \\
YOLOX & OC-SORT & \textit{37.19} & \underline{15.96} & \textit{86.76} & \textit{89.73} & 8,785 & & 313 & \\
Deformable DETR & OC-SORT & \textbf{43.74} & \textbf{22.87} & 83.68 & 88.90 & 5,998 & & 201 & \\
\end{tabular}\\
\bottomrule
\end{tabular}
}
\end{table*}

\begin{table*}[!t]
\centering
\caption{Tracking performance across different detectors and trackers for selected categories on video 2 test set (part 2)}
\label{tab:all_categories_part_2}
\resizebox{\textwidth}{!}{%
\begin{tabular}{c}
\toprule
\begin{tabular}{l l c c c c r r r r}
\multicolumn{9}{c}{
    \raisebox{-0.15\height}{\includegraphics[width=0.4cm]{icons/3_walkerWalking.pdf}}
    WalkerWalking
} \\
\toprule
\multirow{2}{*}{\textbf{Detector}} & \multirow{2}{*}{\textbf{Method}} & \multicolumn{4}{c}{\textbf{Tracking Metrics (\%)}} & \multicolumn{2}{c}{\textbf{Detection Counts}} & \multicolumn{2}{c}{\textbf{Track ID Counts} \vspace{1mm}} \\ \cmidrule(lr){3-6} \cmidrule(lr){7-8} \cmidrule(lr){9-10} &  & \textbf{HOTA} & \textbf{DetA} & \textbf{AssA} & \textbf{LocA} & \textbf{Predictions} & \textbf{GT} & \textbf{Predictions} & \textbf{GT} \\
\midrule
CenterNet & BOT-SORT & 20.84 & 23.22 & 18.87 & 81.35 & 19,178 & \multirow{15}{*}{7,390} & 1,182 & \multirow{15}{*}{5} \\
CenterNet & ByteTrack & 21.49 & 24.52 & 18.99 & 81.44 & 17,947 & & 570 & \\
Deformable DETR & OC-SORT & 24.84 & \textit{38.82} & 15.96 & 80.66 & 8,357 & & 223 & \\
Deformable DETR & BOT-SORT & 26.09 & 32.97 & 20.71 & 81.83 & 12,291 & & 760 & \\
CenterNet & OC-SORT & 26.17 & 31.53 & 21.93 & \textbf{85.02} & 11,263 & & 480 & \\
Faster R-CNN & BOT-SORT & 26.27 & 24.40 & 28.76 & 80.10 & 20,728 & & 876 & \\
Faster R-CNN & ByteTrack & 26.97 & 25.39 & 29.20 & 83.80 & 19,858 & & 470 & \\
RT-DETR & BOT-SORT & 28.66 & 31.03 & 26.73 & 83.25 & 15,267 & & 763 & \\
Faster R-CNN & OC-SORT & 29.27 & 30.08 & 28.99 & 82.06 & 15,450 & & 395 & \\
RT-DETR & ByteTrack & 30.06 & 31.78 & 28.70 & \textit{84.86} & 14,455 & & 455 & \\
YOLOX & BOT-SORT & 32.02 & 31.24 & \underline{33.00} & 81.56 & 16,678 & & 765 & \\
RT-DETR & OC-SORT & 32.35 & \underline{38.13} & 27.75 & 82.39 & 10,653 & & 324 & \\
Deformable DETR & ByteTrack & \underline{32.80} & 34.47 & 31.31 & 82.44 & 11,617 & & 379 & \\
YOLOX & ByteTrack & \textit{32.84} & 32.67 & \textit{33.20} & \underline{83.66} & 15,908 & & 411 & \\
YOLOX & OC-SORT & \textbf{37.71} & \textbf{38.69} & \textbf{36.88} & 81.51 & 12,206 & & 333 & \\
\bottomrule
\end{tabular}\\
\begin{tabular}{l l c c c c r r r r}
\multicolumn{9}{c}{
    \raisebox{-0.15\height}{\includegraphics[width=0.4cm]{icons/4_walkerResting.pdf}}
    WalkerResting
} \\
\toprule
\multirow{2}{*}{\textbf{Detector}} & \multirow{2}{*}{\textbf{Method}} & \multicolumn{4}{c}{\textbf{Tracking Metrics (\%)}} & \multicolumn{2}{c}{\textbf{Detection Counts}} & \multicolumn{2}{c}{\textbf{Track ID Counts} \vspace{1mm}} \\ \cmidrule(lr){3-6} \cmidrule(lr){7-8} \cmidrule(lr){9-10} &  & \textbf{HOTA} & \textbf{DetA} & \textbf{AssA} & \textbf{LocA} & \textbf{Predictions} & \textbf{GT} & \textbf{Predictions} & \textbf{GT} \\
\midrule
RT-DETR & OC-SORT & 0.00 & 0.00 & 83.29 & 0.00 & 4,620 & \multirow{15}{*}{749} & 305 & \multirow{15}{*}{1} \\
CenterNet & OC-SORT & 0.00 & 0.00 & 83.68 & 0.00 & 6,797 & & 459 & \\
RT-DETR & ByteTrack & 0.06 & 0.02 & 84.80 & 0.22 & 8,061 & & 434 & \\
RT-DETR & BOT-SORT & 0.07 & 0.03 & 84.53 & 0.20 & 8,836 & & 732 & \\
CenterNet & ByteTrack & 0.09 & 0.03 & 84.08 & 0.28 & 13,172 & & 551 & \\
CenterNet & BOT-SORT & 0.13 & 0.05 & 82.32 & 0.35 & 14,399 & & 1,150 & \\
Faster R-CNN & OC-SORT & 0.25 & 0.13 & 82.61 & 0.49 & 9,987 & & 389 & \\
YOLOX & OC-SORT & 0.41 & 0.18 & 82.71 & 0.93 & 5,705 & & 307 & \\
Faster R-CNN & BOT-SORT & 0.48 & 0.27 & 82.60 & 0.84 & 15,096 & & 857 & \\
Faster R-CNN & ByteTrack & 0.48 & 0.25 & 82.51 & 0.92 & 14,238 & & 470 & \\
Deformable DETR & OC-SORT & 0.49 & \underline{0.31} & \underline{84.60} & 0.79 & 2,595 & & 184 & \\
YOLOX & BOT-SORT & 0.57 & 0.24 & 82.84 & 1.36 & 9,784 & & 715 & \\
YOLOX & ByteTrack & \underline{0.60} & 0.26 & \textit{84.82} & \underline{1.41} & 9,034 & & 387 & \\
Deformable DETR & BOT-SORT & \textit{1.52} & \textit{0.89} & \textbf{84.98} & \textit{2.60} & 5,592 & & 629 & \\
Deformable DETR & ByteTrack & \textbf{1.58} & \textbf{0.90} & 84.49 & \textbf{2.77} & 5,008 & & 357 & \\
\end{tabular}\\
\bottomrule
\begin{tabular}{l l c c c c r r r r}
\multicolumn{9}{c}{
    \raisebox{-0.15\height}{\includegraphics[width=0.4cm]{icons/6_pushEmptyWheelchair.pdf}}
    PushEmptyWheelchair
} \\
\toprule
\multirow{2}{*}{\textbf{Detector}} & \multirow{2}{*}{\textbf{Method}} & \multicolumn{4}{c}{\textbf{Tracking Metrics (\%)}} & \multicolumn{2}{c}{\textbf{Detection Counts}} & \multicolumn{2}{c}{\textbf{Track ID Counts} \vspace{1mm}} \\ \cmidrule(lr){3-6} \cmidrule(lr){7-8} \cmidrule(lr){9-10} &  & \textbf{HOTA} & \textbf{DetA} & \textbf{AssA} & \textbf{LocA} & \textbf{Predictions} & \textbf{GT} & \textbf{Predictions} & \textbf{GT} \\
\midrule
RT-DETR & BOT-SORT & 9.86 & 3.79 & 81.35 & 26.03 & 9,194 & \multirow{15}{*}{801} & 731 & \multirow{15}{*}{1} \\
CenterNet & OC-SORT & 4.69 & 3.30 & 81.44 & 6.73 & 6,919 & & 460 & \\
Deformable DETR & OC-SORT & \textbf{16.45} & \textbf{9.58} & 80.66 & 28.68 & 3,438 & & 184 & \\
Deformable DETR & ByteTrack & \textit{14.04} & \textit{6.78} & 81.83 & 29.64 & 5,917 & & 354 & \\
Faster R-CNN & ByteTrack & \underline{13.69} & 3.96 & \textbf{85.02} & 47.49 & 14,698 & & 466 & \\
Deformable DETR & BOT-SORT & 13.09 & \underline{6.15} & 80.10 & 28.43 & 6,503 & & 624 & \\
Faster R-CNN & BOT-SORT & 13.05 & 3.69 & 83.80 & \textbf{46.34} & 15,567 & & 854 & \\
Faster R-CNN & OC-SORT & 12.33 & 4.90 & 83.25 & 31.16 & 10,419 & & 388 & \\
RT-DETR & OC-SORT & 11.16 & 5.27 & 82.06 & 23.96 & 4,976 & & 306 & \\
CenterNet & ByteTrack & 10.92 & 3.79 & \textit{84.86} & \textit{31.64} & 13,469 & & 551 & \\
YOLOX & OC-SORT & 10.88 & 4.96 & 81.56 & 24.17 & 6,066 & & 307 & \\
YOLOX & ByteTrack & 10.70 & 4.19 & 82.39 & 27.78 & 9,451 & & 388 & \\
RT-DETR & ByteTrack & 10.51 & 4.12 & 82.44 & 27.22 & 8,386 & & 434 & \\
CenterNet & BOT-SORT & 10.23 & 3.42 & \underline{83.66} & \underline{30.70} & 14,664 & & 1,150 & \\
YOLOX & BOT-SORT & 10.06 & 3.89 & 81.51 & 26.34 & 10,196 & & 717 & \\
\end{tabular}\\
\bottomrule
\end{tabular}
}
\end{table*}

\begin{table*}[!t]
\centering
\caption{Tracking performance across different detectors and trackers for selected categories on video 2 test set (part 3)}
\label{tab:all_categories_part_3}
\resizebox{\textwidth}{!}{%
\begin{tabular}{c}
\toprule
\begin{tabular}{l l c c c c c r r r r}
\multicolumn{9}{c}{
    \raisebox{-0.15\height}{\includegraphics[width=0.4cm]{icons/7_WheelchairGroup.pdf}}
    WheelchairGroup
} \\
\toprule
\multirow{2}{*}{\textbf{Detector}} & \multirow{2}{*}{\textbf{Method}} & \multicolumn{4}{c}{\textbf{Tracking Metrics (\%)}} & \multicolumn{2}{c}{\textbf{Detection Counts}} & \multicolumn{2}{c}{\textbf{Track ID Counts} \vspace{1mm}} \\ \cmidrule(lr){3-6} \cmidrule(lr){7-8} \cmidrule(lr){9-10} &  & \textbf{HOTA} & \textbf{DetA} & \textbf{AssA} & \textbf{LocA} & \textbf{Predictions} & \textbf{GT} & \textbf{Predictions} & \textbf{GT} \\
\midrule
CenterNet & ByteTrack & 26.50 & 33.94 & 20.69 & \textit{88.28} & 23,035 & \multirow{15}{*}{9,672} & 564 & \multirow{15}{*}{5} \\
CenterNet & BOT-SORT & 26.64 & 32.20 & 22.04 & 87.80 & 24,156 & & 1,224 & \\
RT-DETR & BOT-SORT & 27.95 & 36.58 & 21.37 & 87.02 & 16,572 & & 756 & \\
RT-DETR & ByteTrack & 28.76 & 38.32 & 21.61 & 87.55 & 15,754 & & 452 & \\
CenterNet & OC-SORT & 28.88 & 44.25 & 18.87 & 87.84 & 16,363 & & 482 & \\
RT-DETR & OC-SORT & 32.38 & 44.27 & 23.69 & 87.32 & 11,964 & & 329 & \\
Deformable DETR & OC-SORT & 33.22 & \textit{51.20} & 21.59 & 87.76 & 9,871 & & 195 & \\
Deformable DETR & BOT-SORT & 38.80 & 44.61 & 33.77 & 87.29 & 13,365 & & 646 & \\
Faster R-CNN & BOT-SORT & 40.13 & 32.84 & 49.04 & 87.92 & 25,414 & & 874 & \\
Deformable DETR & ByteTrack & 40.39 & \underline{46.60} & 35.02 & 87.93 & 12,768 & & 366 & \\
Faster R-CNN & ByteTrack & 40.99 & 34.15 & 49.19 & \textbf{88.37} & 24,582 & & 475 & \\
Faster R-CNN & OC-SORT & 50.91 & 40.98 & 63.25 & 87.95 & 20,121 & & 391 & \\
YOLOX & BOT-SORT & \underline{51.89} & 41.76 & \textit{64.53} & 87.57 & 19,493 & & 732 & \\
YOLOX & OC-SORT & \textit{57.67} & \textbf{52.02} & \textbf{64.01} & 87.64 & 15,239 & & 314 & \\
YOLOX & ByteTrack & \textbf{60.65} & 43.87 & \textbf{83.85} & \underline{88.14} & 18,700 & & 391 & \\
\bottomrule
\end{tabular}\\
\begin{tabular}{l l c c c c r r r r}
\multicolumn{9}{c}{
    \raisebox{-0.15\height}{\includegraphics[width=0.4cm]{icons/8_WheelchairPusher.pdf}}
    WheelchairPusher
} \\
\toprule
\multirow{2}{*}{\textbf{Detector}} & \multirow{2}{*}{\textbf{Method}} & \multicolumn{4}{c}{\textbf{Tracking Metrics (\%)}} & \multicolumn{2}{c}{\textbf{Detection Counts}} & \multicolumn{2}{c}{\textbf{Track ID Counts} \vspace{1mm}} \\ \cmidrule(lr){3-6} \cmidrule(lr){7-8} \cmidrule(lr){9-10} &  & \textbf{HOTA} & \textbf{DetA} & \textbf{AssA} & \textbf{LocA} & \textbf{Predictions} & \textbf{GT} & \textbf{Predictions} & \textbf{GT} \\
\midrule
CenterNet & OC-SORT & 26.41 & 41.01 & 17.01 & 83.21 & 16,259 & \multirow{15}{*}{9,672} & 475 & \multirow{15}{*}{5} \\
RT-DETR & BOT-SORT & 26.46 & 32.89 & 21.31 & 82.65 & 15,978 & & 767 & \\
RT-DETR & ByteTrack & 27.04 & 34.09 & 21.47 & 82.76 & 15,207 & & 449 & \\
CenterNet & BOT-SORT & 27.16 & 29.59 & 24.94 & 83.17 & 24,449 & & 1,239 & \\
Deformable DETR & OC-SORT & 27.20 & \textit{46.06} & 16.08 & \textit{84.19} & 9,338 & & 190 & \\
RT-DETR & OC-SORT & 28.29 & 39.19 & 20.44 & 82.96 & 11,422 & & 312 & \\
Faster R-CNN & BOT-SORT & 28.48 & 30.25 & 26.83 & 83.04 & 24,948 & & 888 & \\
CenterNet & ByteTrack & 29.50 & 31.15 & 27.94 & 83.27 & 23,161 & & 570 & \\
Faster R-CNN & OC-SORT & 30.46 & 37.28 & 24.89 & 83.06 & 19,678 & & 398 & \\
Faster R-CNN & ByteTrack & 31.17 & 31.41 & 30.93 & 83.27 & 24,070 & & 477 & \\
Deformable DETR & BOT-SORT & 33.88 & 41.22 & 27.86 & 83.69 & 12,671 & & 636 & \\
Deformable DETR & ByteTrack & 34.58 & \underline{42.63} & 28.06 & 84.01 & 12,093 & & 359 & \\
YOLOX & BOT-SORT & \underline{38.34} & 39.08 & \textbf{37.65} & \underline{84.16} & 18,167 & & 727 & \\
YOLOX & ByteTrack & \textit{38.73} & 40.66 & \textit{36.92} & \underline{84.16} & 17,440 & & 391 & \\
YOLOX & OC-SORT & \textbf{41.44} & \textbf{47.40} & \underline{36.27} & \textbf{84.33} & 14,100 & & 318 & \\
\end{tabular}\\
\bottomrule
\begin{tabular}{l l c c c c r r r r}
\multicolumn{9}{c}{
    \raisebox{-0.15\height}{\includegraphics[width=0.4cm]{icons/9_WheelchairPushedUser.pdf}}
    WheelchairPushedUser
} \\
\toprule
\multirow{2}{*}{\textbf{Detector}} & \multirow{2}{*}{\textbf{Method}} & \multicolumn{4}{c}{\textbf{Tracking Metrics (\%)}} & \multicolumn{2}{c}{\textbf{Detection Counts}} & \multicolumn{2}{c}{\textbf{Track ID Counts} \vspace{1mm}} \\ \cmidrule(lr){3-6} \cmidrule(lr){7-8} \cmidrule(lr){9-10} &  & \textbf{HOTA} & \textbf{DetA} & \textbf{AssA} & \textbf{LocA} & \textbf{Predictions} & \textbf{GT} & \textbf{Predictions} & \textbf{GT} \\
\midrule
CenterNet & BOT-SORT & 24.87 & 28.97 & 21.49 & 82.16 & 23,723 & \multirow{15}{*}{9,672} & 1,184 & \multirow{15}{*}{5} \\
CenterNet & OC-SORT & 27.85 & 38.88 & 20.11 & 82.02 & 15,183 & & 468 & \\
RT-DETR & BOT-SORT & 27.88 & 34.09 & 23.25 & 82.46 & 15,772 & & 741 & \\
RT-DETR & ByteTrack & 28.50 & 35.43 & 23.43 & 82.61 & 14,989 & & 434 & \\
RT-DETR & OC-SORT & 30.58 & 41.48 & 23.01 & 82.76 & 11,393 & & 307 & \\
CenterNet & ByteTrack & 32.65 & 30.63 & 34.89 & 82.26 & 22,376 & & 556 & \\
Deformable DETR & OC-SORT & 33.88 & \textbf{53.49} & 21.48 & 86.09 & 10,169 & & 184 & \\
Deformable DETR & BOT-SORT & 35.40 & 45.64 & 27.47 & 86.05 & 13,454 & & 633 & \\
Deformable DETR & ByteTrack & 36.07 & \underline{47.25} & 27.55 & 86.23 & 12,907 & & 357 & \\
Faster R-CNN & BOT-SORT & 42.47 & 30.87 & 58.45 & 82.80 & 24,776 & & 863 & \\
Faster R-CNN & ByteTrack & 43.28 & 31.99 & 58.55 & 82.86 & 23,905 & & 470 & \\
Faster R-CNN & OC-SORT & 45.54 & 38.25 & 54.21 & 82.80 & 19,635 & & 393 & \\
YOLOX & BOT-SORT & \underline{59.05} & 42.74 & \textit{81.59} & \underline{87.06} & 18,740 & & 718 & \\
YOLOX & ByteTrack & \textit{60.16} & 44.29 & \textbf{81.72} & \textit{87.13} & 18,071 & & 387 & \\
YOLOX & OC-SORT & \textbf{65.52} & \textit{53.06} & \underline{80.90} & \textbf{87.14} & 14,659 & & 311 & \\
\end{tabular}\\
\bottomrule
\end{tabular}
}
\end{table*}

\begin{table*}[!t]
\centering
\caption{Tracking performance across different detectors and trackers for selected categories on video 3 test set (Part 1)}
\label{tab:all_categories_video_3}
\resizebox{\textwidth}{!}{%
\begin{tabular}{c}
\toprule
\begin{tabular}{l l c c c c r r r r}
\multicolumn{9}{c}{
    \raisebox{-0.15\height}{\includegraphics[width=0.4cm]{icons/1_ped.pdf}}
    Ped
} \\
\toprule
\multirow{2}{*}{\textbf{Detector}} & \multirow{2}{*}{\textbf{Method}} & \multicolumn{4}{c}{\textbf{Tracking Metrics (\%)}} & \multicolumn{2}{c}{\textbf{Detection Counts}} & \multicolumn{2}{c}{\textbf{Track ID Counts} \vspace{1mm}} \\ \cmidrule(lr){3-6} \cmidrule(lr){7-8} \cmidrule(lr){9-10} &  & \textbf{HOTA} & \textbf{DetA} & \textbf{AssA} & \textbf{LocA} & \textbf{Predictions} & \textbf{GT} & \textbf{Predictions} & \textbf{GT} \\
\midrule
CenterNet & OC-SORT & 52.11 & 59.83 & 45.42 & 85.83 & 5,198 & \multirow{15}{*}{3,995} & 80 & \multirow{15}{*}{5} \\
Faster R-CNN & OC-SORT & 52.47 & 62.94 & 43.78 & 85.33 & 4,975 & & 53 & \\
CenterNet & BOT-SORT & 61.33 & 49.43 & 76.22 & 86.02 & 6,686 & & 237 & \\
RT-DETR & OC-SORT & 61.36 & 71.75 & 52.66 & 85.99 & 4,501 & & 36 & \\
CenterNet & ByteTrack & 61.91 & 51.32 & 74.82 & \underline{86.05} & 6,436 & & 103 & \\
Faster R-CNN & BOT-SORT & 63.66 & 58.37 & 69.48 & 85.38 & 5,589 & & 164 & \\
Faster R-CNN & ByteTrack & 68.18 & 59.93 & 77.65 & 85.48 & 5,444 & & 83 & \\
YOLOX & BOT-SORT & 73.25 & 67.84 & 79.35 & 85.24 & 4,735 & & 130 & \\
YOLOX & ByteTrack & 73.41 & 69.96 & 77.35 & 85.29 & 4,591 & & 71 & \\
Deformable DETR & OC-SORT & 74.47 & \textbf{79.63} & 69.79 & 86.49 & 4,047 & & 17 & \\
YOLOX & OC-SORT & 75.04 & 73.27 & 77.17 & 85.23 & 4,315 & & 53 & \\
RT-DETR & BOT-SORT & 75.91 & 68.16 & \underline{85.12} & 86.01 & 4,796 & & 82 & \\
RT-DETR & ByteTrack & \underline{76.58} & 69.43 & 85.01 & 86.04 & 4,706 & & 40 & \\
Deformable DETR & BOT-SORT & \textit{81.99} & \underline{78.79} & \textit{85.48} & \textit{86.56} & 4,201 & & 47 & \\
Deformable DETR & ByteTrack & \textbf{82.44} & \textit{79.46} & \textbf{85.70} & \textbf{86.70} & 4,166 & & 23 & \\
\bottomrule
\end{tabular}\\
\begin{tabular}{l l c c c c r r r r}
\multicolumn{9}{c}{
    \raisebox{-0.15\height}{\includegraphics[width=0.4cm]{icons/2_cane.pdf}}
    Cane
} \\
\toprule
\multirow{2}{*}{\textbf{Detector}} & \multirow{2}{*}{\textbf{Method}} & \multicolumn{4}{c}{\textbf{Tracking Metrics (\%)}} & \multicolumn{2}{c}{\textbf{Detection Counts}} & \multicolumn{2}{c}{\textbf{Track ID Counts} \vspace{1mm}} \\ \cmidrule(lr){3-6} \cmidrule(lr){7-8} \cmidrule(lr){9-10} &  & \textbf{HOTA} & \textbf{DetA} & \textbf{AssA} & \textbf{LocA} & \textbf{Predictions} & \textbf{GT} & \textbf{Predictions} & \textbf{GT} \\
\midrule
CenterNet & BOT-SORT & 29.90 & 27.51 & 32.54 & 82.02 & 4,222 & \multirow{15}{*}{1,598} & 242 & \multirow{15}{*}{2} \\
Faster R-CNN & BOT-SORT & 34.51 & 37.54 & 31.91 & 82.28 & 3,188 & & 172 & \\
CenterNet & ByteTrack & 35.25 & 28.95 & 43.10 & 81.87 & 3,993 & & 105 & \\
Faster R-CNN & ByteTrack & 35.41 & 39.37 & 32.02 & 82.28 & 3,028 & & 81 & \\
CenterNet & OC-SORT & 40.17 & 36.36 & 44.53 & 81.99 & 2,864 & & 84 & \\
YOLOX & ByteTrack & 44.17 & 50.85 & 38.58 & 81.67 & 2,193 & & 66 & \\
YOLOX & BOT-SORT & 45.04 & 48.47 & 41.98 & 81.66 & 2,313 & & 133 & \\
YOLOX & OC-SORT & 45.99 & 55.91 & 38.32 & 81.71 & 1,916 & & 53 & \\
Faster R-CNN & OC-SORT & 50.60 & 42.67 & \underline{60.05} & 81.88 & 2,587 & & 56 & \\
RT-DETR & OC-SORT & 53.44 & 54.64 & 52.30 & 81.67 & 2,065 & & 34 & \\
Deformable DETR & OC-SORT & 56.33 & \textit{64.44} & 49.39 & \textbf{82.94} & 1,532 & & 21 & \\
Deformable DETR & BOT-SORT & 58.15 & \underline{63.62} & 53.23 & \underline{82.60} & 1,726 & & 48 & \\
Deformable DETR & ByteTrack & \underline{58.66} & \textbf{64.58} & 53.36 & \textit{82.63} & 1,695 & & 20 & \\
RT-DETR & BOT-SORT & \textit{60.08} & 49.76 & \textit{72.56} & 81.69 & 2,373 & & 79 & \\
RT-DETR & ByteTrack & \textbf{61.37} & 51.77 & \textbf{72.74} & 81.71 & 2,278 & & 38 & \\
\end{tabular}\\
\bottomrule
\begin{tabular}{l l c c c c r r r r}
\multicolumn{9}{c}{
    \raisebox{-0.15\height}{\includegraphics[width=0.4cm]{icons/5_wheelchair.pdf}}
    Wheelchair
} \\
\toprule
\multirow{2}{*}{\textbf{Detector}} & \multirow{2}{*}{\textbf{Method}} & \multicolumn{4}{c}{\textbf{Tracking Metrics (\%)}} & \multicolumn{2}{c}{\textbf{Detection Counts}} & \multicolumn{2}{c}{\textbf{Track ID Counts} \vspace{1mm}} \\ \cmidrule(lr){3-6} \cmidrule(lr){7-8} \cmidrule(lr){9-10} &  & \textbf{HOTA} & \textbf{DetA} & \textbf{AssA} & \textbf{LocA} & \textbf{Predictions} & \textbf{GT} & \textbf{Predictions} & \textbf{GT} \\
\midrule
CenterNet         & BOT-SORT     & 38.59  & 18.46  & 80.71  & 85.66  & 3,578   & \multirow{15}{*}{799} & 231  & \multirow{15}{*}{1} \\
CenterNet         & ByteTrack   & 40.13  & 19.88  & 81.06  & 85.93  & 3,312   &                       & 102  &                     \\
Faster R-CNN      & BOT-SORT     & 47.57  & 27.28  & 83.00  & 86.15  & 2,435   &                       & 161  &                     \\
Faster R-CNN      & ByteTrack   & 49.09  & 29.06  & 82.97  & 86.07  & 2,283   &                       & 80   &                     \\
CenterNet         & OC-SORT      & 49.12  & 29.64  & 81.42  & 85.76  & 2,184   &                       & 78   &                     \\
Faster R-CNN      & OC-SORT      & 53.99  & 35.14  & 82.99  & 85.89  & 1,884   &                       & 52   &                     \\
YOLOX             & BOT-SORT     & 62.45  & 44.71  & 87.25  & 90.63  & 1,537   &                       & 121  &                     \\
RT-DETR           & BOT-SORT     & 62.97  & 44.50  & 89.11  & 90.21  & 1,616   &                       & 73   &                     \\
RT-DETR           & ByteTrack   & 64.63  & 46.95  & 88.99  & 90.18  & 1,529   &                       & 36   &                     \\
YOLOX             & ByteTrack   & 65.26  & 48.78  & 87.32  & 90.66  & 1,406   &                       & 67   &                     \\
RT-DETR           & OC-SORT      & 68.84  & 53.22  & 89.05  & 90.22  & 1,347   &                       & 31   &                     \\
YOLOX             & OC-SORT      & 71.59  & 58.73  & 87.26  & 90.63  & 1,160   &                       & 50   &                     \\
Deformable DETR   & BOT-SORT     & \underline{78.31}  & \underline{68.79}  & \underline{89.15}  & \underline{91.45}  & 1,023   &                       & 42   &                     \\
Deformable DETR   & ByteTrack   & \textit{79.76}  & \textit{71.02}  & \textbf{89.59}  & \textbf{91.66}  & 992     &                       & 20   &                     \\
Deformable DETR   & OC-SORT      & \textbf{82.75}  & \textbf{76.64 } & \textit{89.34}  & \textit{91.51}  & 913     &                       & 15   &                     \\
\end{tabular}\\
\bottomrule
\end{tabular}
}
\end{table*}

\begin{table*}[!t]
\centering
\caption{Tracking performance across different detectors and trackers for selected categories on video 3 test set (part 2)}
\label{tab:all_categories_video_3_part_2}
\resizebox{\textwidth}{!}{%
\begin{tabular}{c}
\toprule
\begin{tabular}{l l c c c c r r r r}
\multicolumn{9}{c}{
    \raisebox{-0.15\height}{\includegraphics[width=0.4cm]{icons/3_walkerWalking.pdf}}
    WalkerWalking
} \\
\toprule
\multirow{2}{*}{\textbf{Detector}} & \multirow{2}{*}{\textbf{Method}} & \multicolumn{4}{c}{\textbf{Tracking Metrics (\%)}} & \multicolumn{2}{c}{\textbf{Detection Counts}} & \multicolumn{2}{c}{\textbf{Track ID Counts} \vspace{1mm}} \\ \cmidrule(lr){3-6} \cmidrule(lr){7-8} \cmidrule(lr){9-10} &  & \textbf{HOTA} & \textbf{DetA} & \textbf{AssA} & \textbf{LocA} & \textbf{Predictions} & \textbf{GT} & \textbf{Predictions} & \textbf{GT} \\
\midrule
CenterNet         & ByteTrack   & 1.17   & 0.38   & 3.58   & 82.30  & 3,127   & \multirow{15}{*}{16} & 102  & \multirow{15}{*}{1} \\
CenterNet         & BOT-SORT     & 1.17   & 0.36   & 3.78   & 84.41  & 3,443   &                       & 234  &                     \\
CenterNet         & OC-SORT      & 1.30   & 0.61   & 2.80   & 84.77  & 2,092   &                       & 80   &                     \\
Deformable DETR   & BOT-SORT     & 1.51   & 1.28   & 1.78   & 86.32  & 737     &                       & 43   &                     \\
Deformable DETR   & ByteTrack   & 1.57   & \underline{1.36}   & 1.81   & \textit{87.38}  & 707     &                       & 20   &                     \\
Faster R-CNN      & BOT-SORT     & 1.71   & 0.74   & 3.98   & 80.50  & 1,696   &                       & 162  &                     \\
YOLOX             & ByteTrack   & 1.76   & 1.11   & 2.78   & 83.71  & 1,106   &                       & 69   &                     \\
Faster R-CNN      & ByteTrack   & 1.80   & 0.81   & 4.00   & 80.76  & 1,541   &                       & 80   &                     \\
RT-DETR           & OC-SORT      & 1.82   & 0.98   & 3.45   & 83.99  & 1,366   &                       & 31   &                     \\
Faster R-CNN      & OC-SORT      & 2.08   & 1.09   & 3.96   & 80.02  & 1,135   &                       & 54   &                     \\
YOLOX             & OC-SORT      & 2.12   & \textit{1.52}   & 2.97   & \textbf{87.40} & 870     &                       & 50   &                     \\
YOLOX             & BOT-SORT     & 2.24   & 1.06   & 4.73   & 86.70  & 1,229   &                       & 123  &                     \\
Deformable DETR   & OC-SORT      & \underline{4.56}   & \textbf{1.53}   & \underline{13.58}  & \underline{86.62}  & 617     &                       & 16   &                     \\
RT-DETR           & BOT-SORT     & \textit{4.82 }  & 0.81   & \textit{29.11}  & 83.98  & 1,641   &                       & 79   &                     \\
RT-DETR           & ByteTrack   & \textbf{5.08}   & 0.89   & \textbf{29.54}  & 83.98  & 1,503   &                       & 41   &                     \\
\bottomrule
\end{tabular}\\
\begin{tabular}{l l c c c c r r r r}
\multicolumn{9}{c}{
    \raisebox{-0.15\height}{\includegraphics[width=0.4cm]{icons/4_walkerResting.pdf}}
    WalkerResting
} \\
\toprule
\multirow{2}{*}{\textbf{Detector}} & \multirow{2}{*}{\textbf{Method}} & \multicolumn{4}{c}{\textbf{Tracking Metrics (\%)}} & \multicolumn{2}{c}{\textbf{Detection Counts}} & \multicolumn{2}{c}{\textbf{Track ID Counts} \vspace{1mm}} \\
\cmidrule(lr){3-6} \cmidrule(lr){7-8} \cmidrule(lr){9-10}
&  & \textbf{HOTA} & \textbf{DetA} & \textbf{AssA} & \textbf{LocA} & \textbf{Predictions} & \textbf{GT} & \textbf{Predictions} & \textbf{GT} \\ 
\midrule
RT-DETR           & OC-SORT      & 26.96   & 36.39   & 20.19   & 80.69  & 1,106   & \multirow{15}{*}{1,582} & 32   & \multirow{15}{*}{2} \\
CenterNet         & BOT-SORT     & 44.62   & 30.32   & 65.96   & 80.56  & 3,641   &                        & 231  &                     \\
CenterNet         & ByteTrack   & 46.04   & 32.05   & 66.40   & 80.73  & 3,420   &                        & 104  &                     \\
Deformable DETR   & OC-SORT      & 46.43   & \textbf{53.39}   & 40.63   & \textbf{88.00}  & 1,110   &                        & 16   &                     \\
RT-DETR           & ByteTrack   & 48.67   & 37.11   & 64.17   & 80.51  & 1,446   &                        & 36   &                     \\
RT-DETR           & BOT-SORT     & 48.79   & 36.95   & 64.76   & 80.38  & 1,485   &                        & 80   &                     \\
YOLOX             & BOT-SORT     & 49.90   & 39.16   & 63.71   & 83.11  & 1,600   &                        & 122  &                     \\
YOLOX             & ByteTrack   & 51.16   & 40.94   & 64.02   & 83.24  & 1,473   &                        & 66   &                     \\
Faster R-CNN      & BOT-SORT     & 51.77   & 38.86   & 69.33   & 85.18  & 3,003   &                        & 161  &                     \\
CenterNet         & OC-SORT      & 52.76   & 47.50   & 59.34   & 80.60  & 2,202   &                        & 79   &                     \\
Faster R-CNN      & ByteTrack   & 52.99   & 40.70   & 69.37   & 85.25  & 2,853   &                        & 80   &                     \\
YOLOX             & OC-SORT      & 53.65   & 43.41   & 66.38   & 83.50  & 1,208   &                        & 50   &                     \\
Faster R-CNN      & OC-SORT      & \underline{56.48}  & 46.13   & \underline{69.48}   & \underline{85.36}  & 2,441   &                        & 52   &                     \\
Deformable DETR   & BOT-SORT     & \textit{60.17}   & \underline{51.14}   & \textbf{71.12}   & \textit{87.85}  & 1,264   &                        & 44   &                     \\
Deformable DETR   & ByteTrack   & \textbf{60.67}   & \textbf{52.05}   & \textit{71.04}   & \textbf{88.00}  & 1,222   &                        & 20   &                     
\end{tabular}\\
\bottomrule
\end{tabular}
}
\end{table*}

\begin{table*}[!t]
\centering
\caption{Tracking performance across different detectors and trackers for selected categories on video 3 test set (part 3)}
\label{tab:all_categories_video_3_part_3}
\resizebox{\textwidth}{!}{%
\begin{tabular}{c}
\toprule
\begin{tabular}{l l c c c c r r r r}
\multicolumn{9}{c}{
    \raisebox{-0.15\height}{\includegraphics[width=0.4cm]{icons/7_WheelchairGroup.pdf}}
    WheelchairGroup
} \\
\toprule
\multirow{2}{*}{\textbf{Detector}} & \multirow{2}{*}{\textbf{Method}} & \multicolumn{4}{c}{\textbf{Tracking Metrics (\%)}} & \multicolumn{2}{c}{\textbf{Detection Counts}} & \multicolumn{2}{c}{\textbf{Track ID Counts} \vspace{1mm}} \\ \cmidrule(lr){3-6} \cmidrule(lr){7-8} \cmidrule(lr){9-10} &  & \textbf{HOTA} & \textbf{DetA} & \textbf{AssA} & \textbf{LocA} & \textbf{Predictions} & \textbf{GT} & \textbf{Predictions} & \textbf{GT} \\
\midrule
CenterNet & ByteTrack & 35.42 & 29.73 & 42.34 & 85.70 & 4,351 & \multirow{15}{*}{1,598} & 102 & \multirow{15}{*}{2} \\
CenterNet & OC-SORT & 36.22 & 40.31 & 32.66 & 86.92 & 3,196 & & 84 & \\
CenterNet & BOT-SORT & 38.02 & 28.52 & 50.70 & 86.72 & 4,660 & & 247 & \\
Faster R-CNN & BOT-SORT & 62.82 & 44.08 & \textit{89.53} & \underline{90.32} & 3,258 & & 163 & \\
Faster R-CNN & ByteTrack & 63.77 & 45.82 & 88.75 & 89.91 & 3,108 & & 82 & \\
Faster R-CNN & OC-SORT & 68.27 & 52.30 & 89.11 & 90.26 & 2,720 & & 55 & \\
YOLOX & BOT-SORT & 71.12 & 57.05 & 88.66 & 89.32 & 2,480 & & 132 & \\
YOLOX & OC-SORT & 72.44 & 66.14 & 79.40 & 89.27 & 2,076 & & 52 & \\
RT-DETR & BOT-SORT & 72.52 & 58.91 & \underline{89.29} & 89.80 & 2,422 & & 75 & \\
YOLOX & ByteTrack & 72.58 & 60.02 & 87.78 & 88.83 & 2,338 & & 72 & \\
RT-DETR & ByteTrack & 73.44 & 60.75 & 88.78 & 89.50 & 2,335 & & 38 & \\
RT-DETR & OC-SORT & 76.72 & 66.01 & 89.17 & 89.81 & 2,151 & & 33 & \\
Deformable DETR & OC-SORT & \underline{77.77} & \textbf{81.83} & 73.98 & \textit{90.41} & 1,709 & & 17 & \\
Deformable DETR & BOT-SORT & \textit{83.38} & \underline{77.58} & \textbf{89.61} & \textbf{90.56} & 1,842 & & 46 & \\
Deformable DETR & ByteTrack & \textbf{83.49} & \textit{78.39} & 88.93 & 90.02 & 1,808 & & 21 & \\
\bottomrule
\end{tabular}\\
\begin{tabular}{l l c c c c r r r r}
\multicolumn{9}{c}{
    \raisebox{-0.15\height}{\includegraphics[width=0.4cm]{icons/8_WheelchairPusher.pdf}}
    WheelchairPusher
} \\
\toprule
\multirow{2}{*}{\textbf{Detector}} & \multirow{2}{*}{\textbf{Method}} & \multicolumn{4}{c}{\textbf{Tracking Metrics (\%)}} & \multicolumn{2}{c}{\textbf{Detection Counts}} & \multicolumn{2}{c}{\textbf{Track ID Counts} \vspace{1mm}} \\ \cmidrule(lr){3-6} \cmidrule(lr){7-8} \cmidrule(lr){9-10} &  & \textbf{HOTA} & \textbf{DetA} & \textbf{AssA} & \textbf{LocA} & \textbf{Predictions} & \textbf{GT} & \textbf{Predictions} & \textbf{GT} \\
\midrule
CenterNet         & BOT-SORT       & 31.90   & 29.48   & 34.53   & 85.47  & 4,466   & \multirow{15}{*}{1,598} & 243   & \multirow{15}{*}{2} \\
CenterNet         & OC-SORT        & 34.66   & 42.94   & 28.00   & 85.47  & 2,920   &                         & 80    &                     \\
Faster R-CNN      & OC-SORT        & 35.08   & 48.03   & 25.69   & \textbf{86.33}  & 2,687   &                         & 53    &                     \\
YOLOX             & BOT-SORT       & 40.05   & 54.77   & 29.33   & 85.80  & 2,372   &                         & 126   &                     \\
YOLOX             & ByteTrack     & 41.23   & 57.88   & 29.40   & 85.80  & 2,240   &                         & 68    &                     \\
Faster R-CNN      & BOT-SORT       & 48.50   & 40.85   & 57.60   & \textit{86.30}  & 3,287   &                         & 163   &                     \\
Faster R-CNN      & ByteTrack     & 49.72   & 42.84   & 57.74   & \textbf{86.33}  & 3,129   &                         & 81    &                     \\
CenterNet         & ByteTrack     & 49.92   & 30.96   & \textbf{80.53}   & 85.46  & 4,242   &                         & 102   &                     \\
RT-DETR           & BOT-SORT       & 55.86   & 54.77   & 57.04   & 85.47  & 2,405   &                         & 77    &                     \\
RT-DETR           & ByteTrack     & 56.87   & 56.69   & 57.11   & 85.49  & 2,323   &                         & 37    &                     \\
RT-DETR           & OC-SORT        & 58.15   & 60.22   & 56.21   & 85.69  & 2,109   &                         & 34    &                     \\
YOLOX             & OC-SORT        & 67.54   & 64.43   & \textit{70.85}   & 85.81  & 1,955   &                         & 53    &                     \\
Deformable DETR   & BOT-SORT       & \underline{69.24}   & \underline{71.19}   & 67.41   & \underline{86.26}  & 1,825   &                         & 45    &                     \\
Deformable DETR   & OC-SORT        & \textit{70.12}   & \textbf{74.09}   & 66.46   & \textbf{86.33}  & 1,664   &                         & 17    &                     \\
Deformable DETR   & ByteTrack     & \textbf{70.54}   & \textit{72.27}   & \underline{68.93}   & 86.08  & 1,786   &                         & 22    &                     
\end{tabular}\\
\bottomrule
\begin{tabular}{l l c c c c r r r r}
\multicolumn{9}{c}{
    \raisebox{-0.15\height}{\includegraphics[width=0.4cm]{icons/9_WheelchairPushedUser.pdf}}
    WheelchairPushedUser
} \\
\toprule
\multirow{2}{*}{\textbf{Detector}} & \multirow{2}{*}{\textbf{Method}} & \multicolumn{4}{c}{\textbf{Tracking Metrics (\%)}} & \multicolumn{2}{c}{\textbf{Detection Counts}} & \multicolumn{2}{c}{\textbf{Track ID Counts} \vspace{1mm}} \\ \cmidrule(lr){3-6} \cmidrule(lr){7-8} \cmidrule(lr){9-10} &  & \textbf{HOTA} & \textbf{DetA} & \textbf{AssA} & \textbf{LocA} & \textbf{Predictions} & \textbf{GT} & \textbf{Predictions} & \textbf{GT} \\
\midrule
CenterNet         & OC-SORT        & 40.23   & 42.93   & 37.78   & 85.85  & 2,925   & \multirow{15}{*}{1,598} & 82    & \multirow{15}{*}{2} \\
CenterNet         & BOT-SORT       & 44.14   & 30.28   & 64.36   & 85.74  & 4,376   &                         & 235   &                     \\
CenterNet         & ByteTrack     & 45.10   & 32.15   & 63.28   & 85.52  & 4,119   &                         & 102   &                     \\
Faster R-CNN      & BOT-SORT       & 59.53   & 41.22   & 86.00   & 87.48  & 3,354   &                         & 161   &                     \\
Faster R-CNN      & ByteTrack     & 60.93   & 43.18   & 85.98   & 87.49  & 3,200   &                         & 80    &                     \\
Faster R-CNN      & OC-SORT        & 64.21   & 48.87   & 84.39   & 87.45  & 2,782   &                         & 53    &                     \\
YOLOX             & OC-SORT        & 67.35   & 68.75   & 66.02   & 87.86  & 2,007   &                         & 52    &                     \\
RT-DETR           & BOT-SORT       & 69.49   & 56.19   & 85.96   & 87.31  & 2,440   &                         & 74    &                     \\
RT-DETR           & ByteTrack     & 70.80   & 58.34   & 85.95   & 87.26  & 2,351   &                         & 38    &                     \\
YOLOX             & BOT-SORT       & 71.14   & 58.00   & \textbf{87.29}   & 87.86  & 2,401   &                         & 121   &                     \\
YOLOX             & ByteTrack     & 73.00   & 61.23   & \textit{87.04}   & 87.77  & 2,268   &                         & 67    &                     \\
RT-DETR           & OC-SORT        & 73.71   & 63.19   & 86.00   & 87.30  & 2,167   &                         & 33    &                     \\
Deformable DETR   & BOT-SORT       & \underline{80.53}   & \underline{74.96}   & \underline{86.52}   & \textbf{88.27}  & 1,837   &                         & 43    &                     \\
Deformable DETR   & ByteTrack     & \textit{81.06}   & \textit{76.11}   & 86.34   & \textit{88.23}  & 1,806   &                         & 21    &                     \\
Deformable DETR   & OC-SORT        & \textbf{82.48}   & \textbf{79.17}   & 85.94   & \underline{88.11}  & 1,719   &                         & 15    &                      \\
\end{tabular}\\
\bottomrule
\end{tabular}
}
\end{table*}

\end{document}